\documentclass[11pt]{article}

\usepackage[utf8]{inputenc}
\usepackage{booktabs}
\usepackage[final]{acl}
\usepackage{amsmath}
\usepackage{amssymb}
\usepackage{float}
\usepackage{graphicx}
\usepackage{subfig}
\usepackage{times}
\usepackage{latexsym}
\usepackage{tabularx}
\usepackage{algorithm}
\usepackage{algpseudocode}
\usepackage[T1]{fontenc}
\usepackage[utf8]{inputenc}

\usepackage{microtype}

\usepackage{inconsolata}

\usepackage{graphicx}

{\ignorespaces}
{\ignorespaces}
{\ignorespaces}
{\ignorespaces}

\usepackage[colorinlistoftodos]{todonotes}
\title{Aligned but Not Partner-Specific: How Multimodal LLM Agents Succeed in Reference Games Without Forming Conceptual Pacts}

\author{
\textbf{Po-Ya Angela Wang}\textsuperscript{1}\thanks{This work was conducted while Po-Ya Angela Wang was a research visitor at the Max Planck Institute for Psycholinguistics.},
\textbf{Chinmaya Mishra}\textsuperscript{2},
\textbf{Asl{\i} {\"O}zy{\"u}rek}\textsuperscript{2,3},
\textbf{Paula Rubio-Fernández}\textsuperscript{2,4},
\textbf{Esam Ghaleb}\textsuperscript{2}
\\[4pt]
\textsuperscript{1}Co-Intelligence Humanities AI Future Lab and Graduate Institute of Linguistics, National Taiwan University\\
\textsuperscript{2}Multimodal Language Department, Max Planck Institute for Psycholinguistics\\
\textsuperscript{3}Donders Institute for Brain, Cognition and Behaviour, Radboud University\\
\textsuperscript{4}Institut Jean Nicod, Paris\\[4pt]
\small \textbf{Correspondence:}
\href{mailto:poyaw@ntu.edu.tw}{poyaw@ntu.edu.tw}
and
\href{mailto:esam.ghaleb@mpi.nl}{esam.ghaleb@mpi.nl}
}

\begin{document}
\maketitle
\begin{abstract}
Repeated reference games test whether interlocutors replace their initially long descriptions with shorter, partner-specific expressions grounded in shared interaction history; that is, with conceptual pacts. Prior work shows that multimodal LLMs fail to become more efficient across rounds, although they align on the labels they use. However, how can we determine whether this alignment reflects partner-specific grounding rather than a shared task vocabulary? We address this by comparing competent multimodal agent dyads with human dyads from the KTH Tangrams corpus. Our novel methodological contribution is a pragmatically constrained pseudo-dyad baseline: rounds from two different real dyads describing the same target at comparable trajectory positions are paired, preserving referential task structure while removing shared partner history. This enables us to test whether the observed label alignment depends on interaction with a specific partner. Across three measures (task competence, description strategy, alignment dynamics), we find clear differences. Humans reduce effort through entrainment, compressing descriptions and increasing label alignment with partners. Agents instead maintain fixed effort levels, producing verbose descriptions from round one, with near-ceiling label overlap that is statistically indistinguishable between real and pseudo dyads. MLLMs thus achieve coordination without conceptual pacts, succeeding by verbose description rather than by forming the compact, history-dependent referring expressions characteristic of human dialogue.

\end{abstract}

\section{Introduction}

When two people play a referential communication game and repeatedly describe a set of abstract shapes, their referring expressions converge: lengthy first-round descriptions (``the one that looks like an amber-orange colored dinosaur'') compress into compact labels (``the dinosaur''). Such labels become conceptual pacts, which are temporary, partner-specific agreements about how to refer to an object, grounded in shared interaction history \citep{clark1986referring,brennan1996conceptual,pickering2004mechanistic}. Decades of work using the tangram naming paradigm have documented systematic shortening across rounds \citep{hawkins2020characterizing} and the generalisation of partner-specific pacts to community norms \citep{hawkins2023partners}.

\begin{figure*}[tbp]
    \centering
    \includegraphics[width=\linewidth]{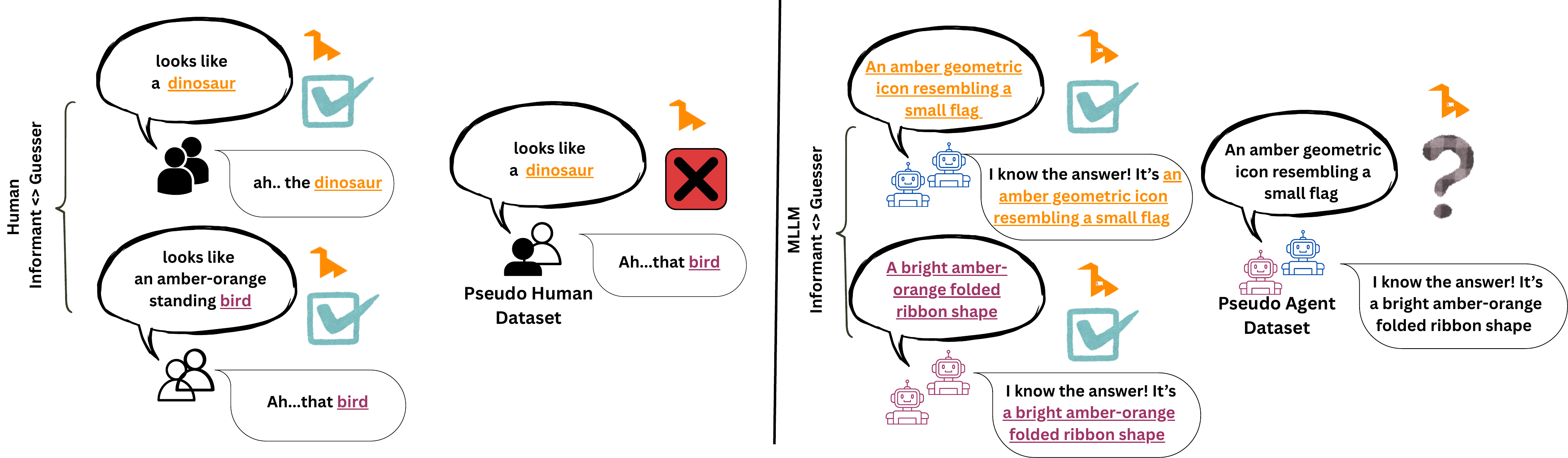}
    \caption{Pseudo-dyad baseline for diagnosing partner-specific grounding and alignment. A pseudo-dyad pairs rounds from two different real dyads describing the same target at comparable trajectory positions: both contributors have prior experience referring to the target, but they do not share that experience. Real human partners (left) converge on shared labels for the target tangram (\checkmark); in the pseudo-paired rounds, the label established in one dyad cannot be retrieved by the contributor from the other (\textbf{X}). The red \textbf{X} marks the absence of shared history, not an inability to communicate. Real agents (right) produce matching verbose descriptions and succeed (\checkmark); a pseudo-paired agent with no shared history (\textbf{?}) still produces near-identical wording.}
    \label{fig:baseline}
\end{figure*}
The progress in multimodal LLMs raises a natural question: when agent dyads are placed in the same cooperative referential task, does their apparent alignment reflect partner-grounded conceptual adaptation or just statistically consistent output shaped by pretrained priors? Existing results are suggestive but leave the question open. \citet{hua2024talk} show that state-of-the-art MLLMs comprehend an interlocutor's increasingly efficient
language, but do not produce shorter messages themselves. \citet{shaikh2024grounding} show that LLMs presume common ground rather than building it incrementally. Crucially, neither study includes a check that removes interaction history while preserving the task, model, and turn structure. Without such a control, it is hard to tell whether the observed lexical alignment is driven by partner-specific grounding or by a shared task vocabulary learned during pretraining.

Rather than testing convention-like compression broadly or explaining its internal cause, we use a pseudo-dyad baseline to ask a more fundamental question: Do MLLM agents use shared interaction history to form partner-specific conceptual pacts in repeated reference games? We address this question through several methodological and empirical contributions:

\begin{itemize}
    \item A case study where we adapt the KTH Tangrams task for MLLM dyads and conduct a comparative analysis of the resulting MLLM dyadic corpus\footnote{The code is publicly available in the repository: \url{https://github.com/diff94/tan_ana}.} against the KTH Tangrams human corpus \citep{shore2018kth}: 45 MLLM dyads / 905~rounds and 42 human dyads / 3,288~rounds, respectively.
    
    \item A pragmatically constrained pseudo-dyad extraction algorithm that pairs rounds across source dyads under matched-target, matched-round, matched-turn-count, and success-parity constraints, providing a pragmatically coherent baseline for separating interaction-specific grounding from global style alignment (Figure~\ref{fig:baseline}). 
    
    \item A comprehensive analysis across (i)~task competence and effort, (ii)~description strategy and lexical composition, and (iii)~ micro- and macro-alignment dynamics, designed so that any pair of dyadic corpora can be compared on the same axes.

\end{itemize}

Our results show that MLLM agent dyads achieve coordination without conceptual pacts: near-ceiling lexical overlap that is statistically indistinguishable between real and pseudo conditions, contrasting with a robust human real--pseudo gap that grows over time and survives dyad-level permutation testing.

\section{Related Work}
\label{sec:related}

\paragraph{Reference games and conceptual pact formation in humans.}

The collaborative theory of reference holds that referring expressions are jointly constructed through grounding acts \citep{clark1986referring,clark1996using}. In the seminal tangram experiments, descriptions shorten as pairs converge on mutually acceptable labels \citep{clark1986referring}. \citet{brennan1996conceptual} formalized
this as conceptual-pact formation.

\citet{hawkins2020characterizing} characterize the convergence as a compression with systematic syntactic structure, where closed-class units drop out, leaving short open-class labels. \citet{hawkins2023partners} model how partner-specific pacts generalize to community norms. Terminologically, a pact is dyad-specific and renegotiable, whereas a convention is a community-level, more durable form--meaning mapping. Theoretically, alignment has been attributed to automatic priming \citep{pickering2004mechanistic}, strategic accommodation \citep{giles1991accommodation,giles2024cat}, and effort--informativeness trade-offs \citep{levshina2021efficiency,levshina2022communicative}.

Crucially for our methodology, \citet{healey2014divergence} and \citet{duran2019align} establish that baselines are essential for distinguishing genuine alignment from chance repetition: lexical overlap arises mechanically whenever two speakers describe the same referents, so any claim about interaction-specific grounding must be made against a counterfactual that holds the task constant while removing the interaction. Our pragmatically constrained pseudo-dyad design (Section~\ref{sec:method-pseudo}) operationalizes this principle for MLLM dyads.

\paragraph{(M)LLMs in referential dialogue.}

Across the (M)LLM literature, lexical alignment is reliably observed but does not suggest grounding. LLM-generated dialogue shows exaggerated, monotonically increasing alignment, unlike humans, whose alignment plateaus and is shaped by the particular partner \citep{mayor2025llm}. Convention formation can be elicited only when training jointly rewards success and message cost \citep{vaduguru2026success}.
Visually grounded alignment shows the same pattern: LVLMs fail to track developing partner-specific conceptual pacts as overhearers \citep{wang2025overhearing}, show lexical-adaptation deficits in self-play \citep{imai2025measuring}, style-match but diverge from humans strategically \citep{kwon2025evaluating}, and fail to establish common ground in factorial human/AI designs \citep{zeng2026lvlmsACL}. Related work on multi-agent LLM cooperation and emergent communication \citep{hu2024survey,chen2024agentverse,
akata2025repeated,wu2024teamup,nath2025collaborate,
lazaridou2016towards,lazaridou2020emergent,dutta2025emergent, kumar2024leets} echoes this gap between coordination and human-style adaptation\footnote{Terminology here is not uniform: several of these studies use convention for what we, following \citet{brennan1996conceptual}, call a conceptual pact. We preserve each work's own framing when referring to it, and reserve the narrower usage for our own claim (Section~\ref{sec:method-terms}).}.

However, alignment is typically computed on real dyads and compared either against a human ceiling \citep{mayor2025llm}, against absent-of-history single-turn prompts \citep{shaikh2024grounding}, or across model families \citep{imai2025measuring}. Prior work lacks a within-condition control that keeps the task, model, prompt, and turn structure constant while removing only the shared interaction history. Without this control, the observed alignment can have two possible explanations. It may reflect partner-specific grounding (the human-like interpretation), or it may result from shared task vocabulary combined with pretrained style (the null interpretation). Existing MLLM studies cannot tell these two explanations apart.

We close this gap by extending the pseudo-dyad methodology of \citet{duran2019align} and \citet{ghalebetal2024} to the multimodal agent setting under four pragmatic constraints, severing partner history while preserving dialogue structure.

\section{Method}
\label{sec:method}
We study alignment in a cooperative referential task using Tangram shapes. Tangrams are visually rich but hard to name in a fixed way, so speakers must invent descriptions (``the one that looks like a bird'') to help their partner identify the target. Repeated references to the same shapes then let us ask whether MLLMs simply reuse descriptions made likely by the task and their pretrained priors. Our pipeline has four steps: (i) use KTH Tangrams as the human baseline, (ii) adapt the task for MLLM dyads, (iii) select a model that performs the task the best, and (iv) build pseudo-dyads that preserve task structure while removing partner history.

\paragraph{Terminology used:}
\label{sec:method-terms}
We define the main terms used throughout the paper to make the task structure, corpus units, and alignment measures easier to follow.
\begin{itemize}
    \item A \emph{turn} is a single message produced by one speaker within a round.
    \item A \emph{round} is a complete exchange ending when a guess is registered (a \emph{trial} in the classic referential-game literature)
    \item A \emph{dyad} is a pair of speakers across all rounds in one session. A dyad's \emph{trajectory} is the ordered sequence of its rounds. A round's \emph{trajectory position} is its index in that sequence. 
    \item A \emph{label violation} is any informant turn in which the two-letter target tag appears verbatim (case-insensitive exact match). 
    \item A \emph{lexical core} is a cluster of near-equivalent shared words.
 
    \item A \emph{view} is an image/display with tangram shapes shown to the MLLM dyads (Section~\ref{sec:method-cores}). 
    \item We reserve \emph{conceptual pact} (\emph{partner-specific pact}) for the transient, flexible, partner-specific expression-referent associations a dyad establishes and may renegotiate during an interaction \citep{brennan1996conceptual}, and \emph{convention} for community-wide, relatively context-independent form--meaning mappings persisting beyond any single pair. All measurements are conducted at the pact level, and we make no claim about community-level convention formation. 
\end{itemize}

\begin{figure*}[ht]
    \centering
    \includegraphics[width=\linewidth]{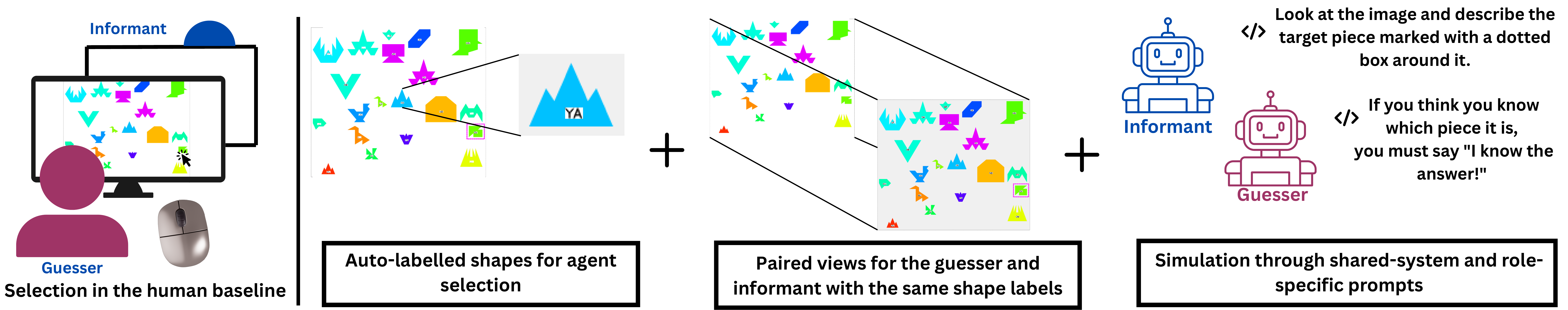}
    \caption{This interactive MLLM adaptation of the KTH Tangrams task replaces human clicking with randomised, two-letter shape tags on paired informant-guesser views, using system prompts to prevent tag leak. It tracks a multimodal memory of utterances and visual configurations across rounds, while managing clarification turns, automated scoring, failure feedback, and role-swapping progression.}    \label{fig:method}
\end{figure*}

\subsection{Human paradigm and KTH baseline}
\label{sec:method-human}
% \amber{I make it shorter}
We follow the human reference-game paradigm of the KTH Tangrams corpus \citep{shore2018kth}, which contains 42 two-party reference game dialogues in English. In each round, an ``informant'' sees a board of 20 tangram pieces (drawn from 17 distinct shape outlines, with some outlines recurring in different colours and/or sizes) and one target highlighted; the ``guesser'' sees the same board without the highlight and must identify the target from the informant's descriptions. A correct selection awards $+1$ points, swaps the roles, and moves only the selected piece to a new random location, leaving other pieces where they were for the next round; incorrect selections cost $-2$ and extend the same round until the target is identified. Every session recycles the same set of shapes, so later mentions are genuine coreferences: a human dyad refers to an individual piece 3.93 times per session on average (median 4, maximum 15; Appendix~\ref{app:coref}). The corpus comprises 3{,}288 rounds (10{,}254 turns).

% \subsection{Terminology}

\subsection{Adapting the paradigm for MLLM dyads}
\label{sec:method-adapt}

Adapting a human reference game into an MLLM-runnable cooperative task required a non-trivial redesign of the testing materials. Four issues had to be solved before we could simulate the tangram task with MLLM dyads, as discussed below. Figure~\ref{fig:method} provides a visual overview of the process. First, human participants click the target shape on a screen, whereas MLLM agents cannot. We must give each shape a machine-readable identifier that the guesser/informant can recognise and produce in text \emph{without leaking} it to one another during interaction. Second, KTH stores the informant view (with a magenta-dashed highlight box around the target) and the guesser view (without the box) as separate screenshots with inconsistent filenames. To run the game, we need each round's two views to be reliably paired. Third, the KTH interface handles role-swaps (between participants being ``guesser'' and ``informant''), piece movement, and scoring automatically. We must reproduce this for MLLM dyads. Fourth, importantly, we maintain a conversation-level memory that dynamically preserves multi-modal context by anchoring both the utterances exchanged and the specific visual configurations of the Tangram boards discussed across rounds.

\paragraph{Auto-labelling the tangram shapes and pairing views.} 
We annotate each shape with a two-letter label that acts as the MLLM equivalent of clicking. Labels had to (a)~be reliably readable by the MLLM,
(b)~not be revealed by the informant to the guesser, and (c)~stay identical in the informant's and guesser's views.  Section~\ref{sec:method-models} addresses (a) and (b), and Appendix~\ref{app:pipeline} addresses (c) with an automatic pairing and labelling pipeline. The 80 informant--guesser view pairs are consecutive screenshots from a single real KTH session (session~26), maintaining the original single-piece movement rather than reshuffling the board between rounds. The tagged final pairs are manually checked.

\paragraph{Game protocol for MLLM dyads} 
A controller\footnote{ To avoid confusion, we referred to the computer system message as the controller in the following discussion.}  simulates gameplay using two prompt types: a \emph{system prompt} giving task-level instructions (setup, goals, rules) and \emph{functional-role prompts} giving role-specific instructions. Full prompts are in Appendix~\ref{app:prompts}. At the start of the round, the controller sends the highlighted view to the informant and the unhighlighted view to the guesser, and issues each agent its role-specific prompt (Figure~\ref{fig:method}). The guesser may ask clarification questions; once it commits, it must say ``I know the answer!'', triggering a system query for the target label. The controller then checks the label, awards $+1$/$-2$, and either continues the round (on failure) or advances with a role swap (on success). Each of the 45 agent dyads sees the same canonical 80-view sequence from a staggered starting offset. Using one human session as the shared visual environment reduces environmental diversity, but holds the visual materials constant and makes exact-piece matching possible when pseudo-dyads are built (Section~\ref{sec:method-pseudo}). Since a session halts as soon as a label violation is detected, dyads finish at varying points, leading to unequal image coverage. Both of these factors are addressed in the Limitations.

\subsection{Model selection}
\label{sec:method-models}
We need the candidate models to accomplish the basic task so that we can investigate the core objective: how humans and machines form partner-specific pacts. We screened against three criteria: (i)~robust visual discrimination of target tangrams, (ii)~strict task compliance (i.e., avoiding explicit target-label leak), and (iii)~support for multimodal interaction history across consecutive rounds. Of three piloted models, \texttt{Llama-3.2-Vision} (latest)  \citep{grattafiori2024llama3} failed ~(i), reaching only 18.75\% accuracy on a tag recognition check and rendering downstream alignment measures uninterpretable. GPT-4o (\texttt{gpt-4o-2024-11-20}) \citep{openai2024gpt4o} was visually accurate (90.0\%), but failed~(ii), frequently leaking the target label during challenging rounds and bypassing the required communicative effort. For GPT-5 (\texttt{gpt-5-2025-08-07}) \citep{openai2025gpt5}, we evaluated 12 configurations across reasoning effort and text verbosity (see Appendix~\ref{app:config} for the full sweep). The selected configuration (high reasoning, medium verbosity) maximised exact-match success (94\%) with 0\% label violations\footnote{ The 0\% here refers to the 36-run pilot sweep; in the full runs the controller still terminates a session on the first label violation, which is why agent sessions are shorter than human ones (Appendix~\ref{app:coref}).}.

Consequently, we use GPT-5 in this configuration for all subsequent experiments. A key consideration was that OpenAI's ``Responses API'' exposes server-side conversation states via \texttt{previous\_response\_id}, providing lossless multimodal interaction history across visual turns. However,  the evaluation protocol remains strictly model-agnostic for future generalisability.

\begin{figure}[t]
    \centering
    \includegraphics[width=1\linewidth]{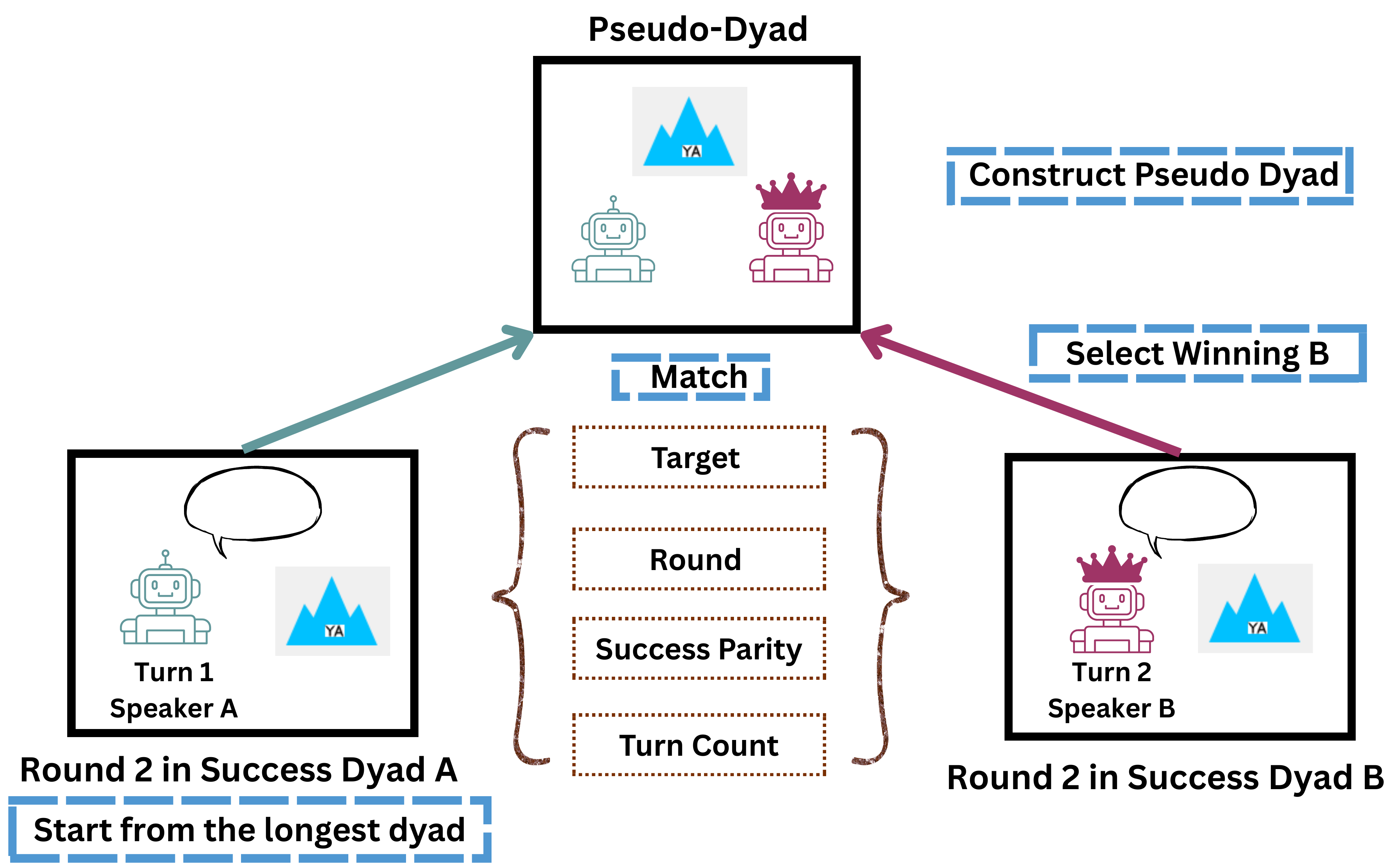}
    \caption{Pseudo-dyad generation. For each round in a source dyad $A$ (left), we search the remaining pool of real dyads for a round $r_b$ in another dyad $B$ (right) that matches on target, round position, turn count, and success outcome. The best-matching pair $(r_a, r_b)$ is locked in as one round of the pseudo-dyad $P_{AB}$ (top). Source dyads are processed in descending trajectory-length order so that long dyads claim partners first.}
    \label{fig:pseudo}
\end{figure}
\subsection{Pragmatically constrained pseudo-dyad generation}
\label{sec:method-pseudo}
\paragraph{Definition and selection of pseudo-dyads.}
A pseudo-dyad pairs rounds from two different real dyads describing the same target at comparable discourse positions in their respective trajectories. Both contributors have prior experience referring to that target, but they do not share that experience. By holding the referential task constant while removing the shared interaction history, this method enables us to determine whether observed alignment reflects partner-specific grounding or only shared task vocabulary plus pretrained style. The procedure has two levels.
\paragraph{Round-level matching.}
For each source round $r_a$ from dyad $A$, a candidate round $r_b$ from dyad $B$ must have the same target identity, the same shape outline for humans and the same piece for agents, and satisfy the tier-specific round-index tolerance. These constraints are never relaxed. The remaining constraints are applied through an ordered fallback cascade. For each $r_a$, the algorithm uses the first fallback stage containing an admissible unused candidate and selects the $r_b$ with the lowest round-level score in that stage (see Appendix~\ref{app:pseudo-scoring}).

\paragraph{Dyad-level pairing.}
Within each tier, real dyads are processed from longest to shortest, so longer dyads claim partners first (Figure~\ref{fig:pseudo}). For each dyad $A$, the algorithm evaluates every remaining candidate $B$ using the greedy round-level procedure and discards candidates that do not
meet the tier's minimum number of matches. It then retains the candidate that maximises the unnormalised aggregate
$S(A,B)=\sum_{(r_a,r_b)}s(r_a,r_b)$. Because this aggregate reflects both the number of matches and the magnitudes of their score components, the selected partner need not provide either the most matches or the smallest average deviation. Each source dyad is paired at most once. The complete score, tier-specific tolerances and weights, and selection procedure appear in Appendix~\ref{app:pseudo-scoring}.

Pseudo-dyads provide a diagnostic check rather than a causal test. If shared interactional history drives alignment, removing that history should weaken alignment when referent identity, trajectory position, and interaction structure are controlled as closely as the matching procedure permits. The procedure yields 20 agent pseudo-dyads containing 238 rounds, constructed from 40 of the 45 source dyads, and 21 human pseudo-dyads containing 204 rounds, constructed from all 42 source dyads.

Only source rounds with valid role identities and at least one retained conversational turn from each role are eligible for pseudo-dyad construction. Consequently, 1{,}016 of the 3{,}288 human rounds ($30.9\%$) are excluded from the pseudo-dyad source pool: each contains an informant turn but no logged guesser identity or guesser turn. These silent-selection rounds remain part of the real human corpus; their exclusion applies only to pseudo-dyad construction. The agent source pool similarly excludes 6 of 905 rounds that end after an informant label violation without a guesser turn. Together with the greater variation in human trajectory length and turn structure, these eligibility restrictions help explain why the larger human corpus yields slightly fewer matched pseudo-rounds.

Every matched pair satisfies its tier's round-index tolerance, and no pair differs by more than two rounds. Round indices match exactly for 82\% of agent pairs and 42\% of human pairs. Both source outcomes are observed and successful for 95.0\% of agent pairs and 99.5\% of human
pairs. The matched rounds cover all 17 human shape outlines and all 20 agent pieces on both sides of the pseudo-dyads.

Conversation-length differences run in opposite directions. Averaging turns per round first within each dyad and then across dyads, agent pseudo-dyads have 14.7\% fewer turns per round than real agent dyads, whereas human pseudo-dyads have 15.0\% more than real human dyads. These opposing shifts do not provide a common length-based explanation for the reported population difference. Full diagnostics appear in Appendix~\ref{app:pseudo-scoring}.
% Conversation-length differences run in opposite directions. Averaging turns per round within each dyad and then across dyads, agent pseudo-dyads have 14.7\% fewer turns per round than real agent dyads, whereas human pseudo-dyads have 15.0\% more, so human pseudo rounds afford more opportunity for overlap yet show far less of it, and agent pseudo rounds afford less yet show no less. Length therefore cannot explain the real--pseudo dissociation. Full diagnostics appear in Appendix~\ref{app:pseudo-scoring}.

\subsection{Lexical alignment extraction}
\label{sec:method-cores}
We operationalize lexical alignment as the cross-speaker reuse of words or multiword sequences within a dyad, detected using the automated procedure introduced by \citet{ghalebetal2024}. We then compute alignment over \emph{lexical cores}: clusters of near-equivalent shared words that preserve the same content-based referential basis while abstracting away from minor morpho-lexical variation, so ``the angular piece'', ``angular shape'', and ``angled form'' are mapped to the same core, \textsc{angular\_shape}. Specifically, we filter out frequent words, function words, and shared surface forms with a content-word ratio below 0.2. Then, we cluster the remainder using a union-find algorithm based on their content-lemma sets, merging forms that share at least one content lemma at Jaccard overlap $\geq 0.6$ (Appendix~\ref{app:cores}). Hence, a lexical core provides an operational approximation to a conceptual pact by grouping near-equivalent surface forms that share the same content-based referential basis.

\paragraph{Dataset Summary}
\label{sec:data}

\begin{table}[t]
\centering
\scriptsize
\setlength{\tabcolsep}{5pt}
\begin{tabular}{lrrrr}
\toprule
& \textbf{A-Real} & \textbf{A-Pseudo} & \textbf{H-Real} & \textbf{H-Pseudo} \\
\midrule
Dyads & 45 & 20 & 42 & 21 \\
Rounds & 905 & 238 & 3,288 & 204 \\
Total turns & 2,039 & 508 & 10,254 & 945 \\
Mean rounds/dyad & 20.11 & 11.90 & 78.29 & 9.71 \\
Mean turns/round & 2.25 & 2.13 & 3.12 & 4.63 \\
Mean turns/dyad & 45.31 & 25.40 & 244.14 & 45.00 \\
Turns-to-success & 2.03 & N/A & 2.56 & N/A \\
Success rate (\%) & 95.03 & N/A & 99.79 & N/A \\
Word tokens & 117,026 & 28,619 & 68,187 & 6,171 \\
Shared surface forms & 5,151 & 815 & 703 & 14 \\
Mean surface forms/dyad & 114.47 & 40.75 & 16.74 & 0.67 \\
Lexical cores & 3,081 & 527 & 503 & 11 \\
Mean cores/dyad & 68.47 & 26.35 & 11.98 & 0.52 \\
\bottomrule
\end{tabular}
\caption{Dataset summary. A = agent; H = human. Turn counts reflect conversational turns only; system and label turns excluded. Pseudo-dyads have no game outcomes.}
\label{tab:dataset}
\end{table}

Table~\ref{tab:dataset} presents corpus-level statistics for the four dyad sets. Two findings are not visible in the table. First, the human pseudo-dyad baseline yielded only 14 shared surface forms, occurring in just 8 of the 21 pseudo-dyads, confirming that alignment in real human dyads reflects genuine interactive convergence rather than task-vocabulary artefacts. Second, agent utterances average $\sim$59 tokens per turn against $\sim$7 for humans, accounting for the nearly six-fold excess in shared lexical cores per dyad despite fewer rounds (full success-only breakdown in Appendix~\ref{app:success_only}).

\section{Results and Discussion}
\label{sec:results}
We evaluate whether task success in MLLM dyads reflects the same mechanism that supports human referential success, i.e., the formation of compact, partner-specific pacts. We first establish whether agents can solve the task at all, then ask whether success is accompanied by declining interactional effort, whether descriptions compress over repeated reference, and finally whether apparent lexical alignment depends on shared partner history under the pragmatically constrained pseudo-dyad control. 

Because agent and human dyads differ in absolute trajectory length
(20.1~vs.~78.3 rounds per dyad on average; Table~\ref{tab:dataset}), several analyses bin each dyad's rounds into ten equal-sized deciles of its own trajectory, i.e., decile~1 is the first 10\% of that dyad's rounds, and decile~10 is the last 10\%. Aggregating across dyads at the decile level puts short and long trajectories on the same temporal scale.

\subsection{Task competence and effort}
\label{sect:task_comp_effort}
\paragraph{Competence.} Both populations solve the task at high rates, but only humans show the signature of accumulating efficiency. Humans succeed on 99.79\% of rounds and agents on 95.03\%. A logistic GLMM over agent rounds (\texttt{is\_success} $\sim$ \texttt{round\_z} + $(1\mid\textsc{dyad})$)
shows that success declines strongly with round number
($\hat{\beta}_{\text{logit}} = -1.11$, 95\% CI $[-1.36, -0.85]$, OR $= 0.33$ $[0.26, 0.43]$): each one-SD increase in round number triples the odds of failure. Agent errors, therefore, concentrate in late rounds, those that draw most on accumulated referential history. 

\begin{figure}[t]
\centering
\includegraphics[width=\columnwidth]{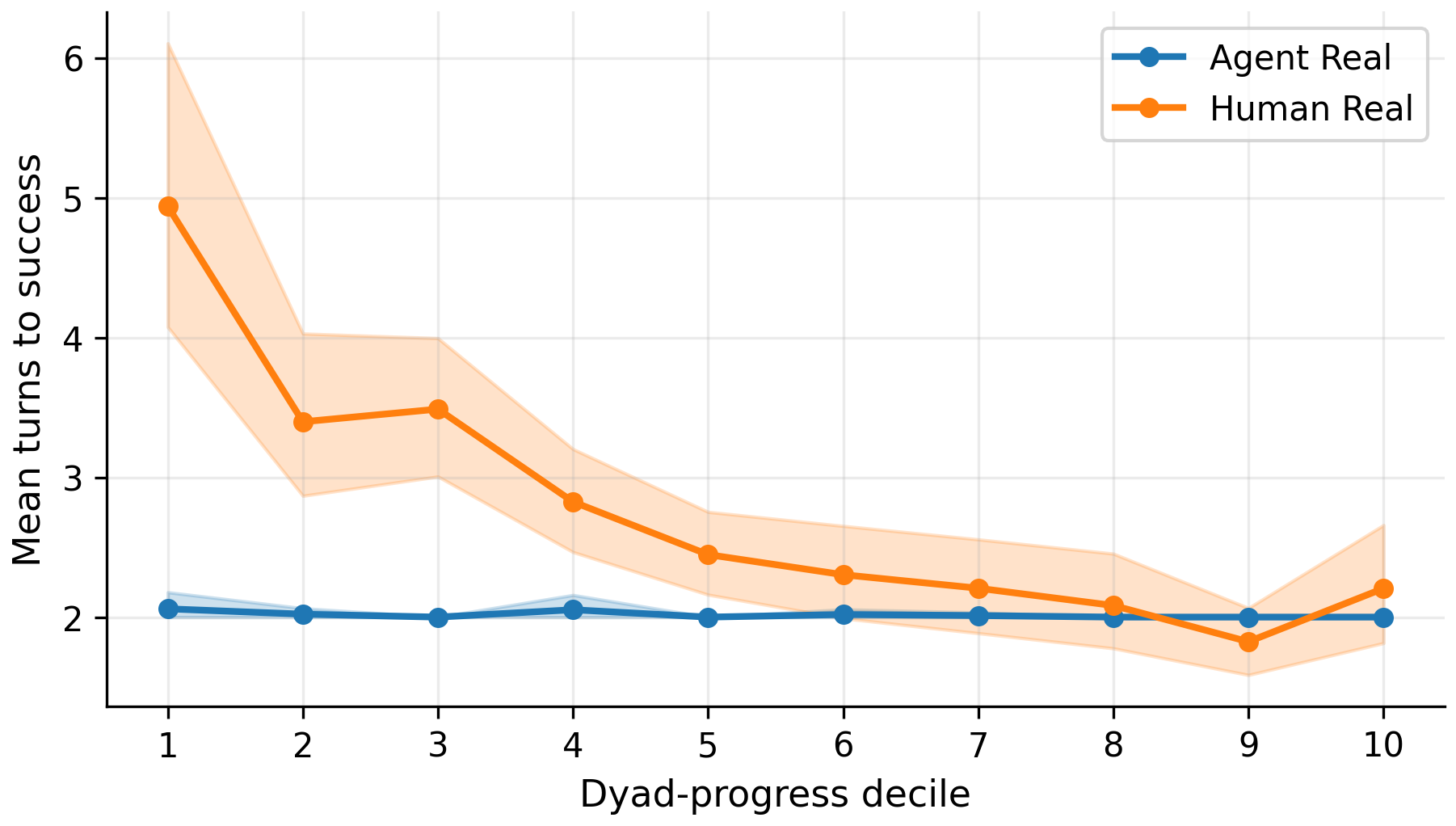}
\caption{Mean turns-to-success across dyad-progress deciles. The $x$-axis is divided into 10 quantile bins per dyad so that human and agent trajectories are placed on a comparable temporal footing.}
\label{fig:turns-to-success}
\end{figure}

\paragraph{Effort.}
Humans average 2.56 conversational turns per correct guess and
agents 2.03, but the trajectories differ sharply. A linear
mixed-effects model on success-only rounds
(\texttt{turns\_to\_success} $\sim$ \texttt{round\_z} +
$(1\mid\textsc{dyad})$) gives a large negative slope for humans
($\hat{\beta} = -0.713$, $p \ll 0.005$, $n = 3{,}281$, 42 dyads) and a near-zero slope for agents ($\hat{\beta} = -0.017$,
$p = 0.163$, n.s., $n = 860$, 45 dyads). Figure~\ref{fig:turns-to-success} visualizes the dissociation on success-only rounds: humans begin near $\sim$5~turns in their earliest deciles and converge toward $\sim$2
by mid-trajectory, while agents remain flat at $\sim$2 throughout.
Failure-inclusive trajectories (Appendix~\ref{app:tso_failure}) show occasional spikes from failed rounds; these are absent from the
success-only view, confirming that failed rounds are substantially longer than successful ones.

Agents thus reach high success rates by exhaustive description rather than by developing efficient conceptual pacts. The cost surfaces in the late-round drop-off, where accumulated referential history matters most. Humans, by contrast, achieve near-perfect success with progressively less effort: their mean turns-to-success falls from 3.93 over the first 20 rounds to 2.56 across the session as a whole (Table~\ref{tab:first20}). This is evidence that MLLM agents fail to benefit from accumulated interaction history at the level of conversational efficiency\footnote{ Throughout, \emph{accumulated interaction history} means what a specific pair has actually exchanged, rather than \emph{common ground}, which also covers what partners share without having interacted.}.

\begin{table}[t]
\centering\small
\setlength{\tabcolsep}{4pt}
\begin{tabular}{lrrrr}
\toprule
& \multicolumn{2}{c}{\textbf{Human}} & \multicolumn{2}{c}{\textbf{Agent}} \\
\cmidrule(lr){2-3}\cmidrule(lr){4-5}
& \textbf{First 20} & \textbf{All} & \textbf{First 20} & \textbf{All} \\
\midrule
Rounds                  & 840 & 3,288 & 648 & 905 \\
Success rate (\%)       & 100.00 & 99.79 & 96.30 & 95.03 \\
Turns-to-success        & 3.93 & 2.56 & 2.03 & 2.03 \\
\bottomrule
\end{tabular}
\caption{Early-window versus whole-session statistics. ``First 20'' is the first 20 rounds of each dyad and is a subset of ``All'', so the two columns are not independent samples. Humans shorten (3.93 $\to$ 2.56 turns); agents do not (2.03 (rounded from 2.0321) $\to$ 2.03 (rounded from 2.0256) turns).}
\label{tab:first20}
\end{table}

\subsection{Description Strategy}
\begin{figure}[tbp]
  \centering
  \hfill
  {\includegraphics[width=\linewidth]{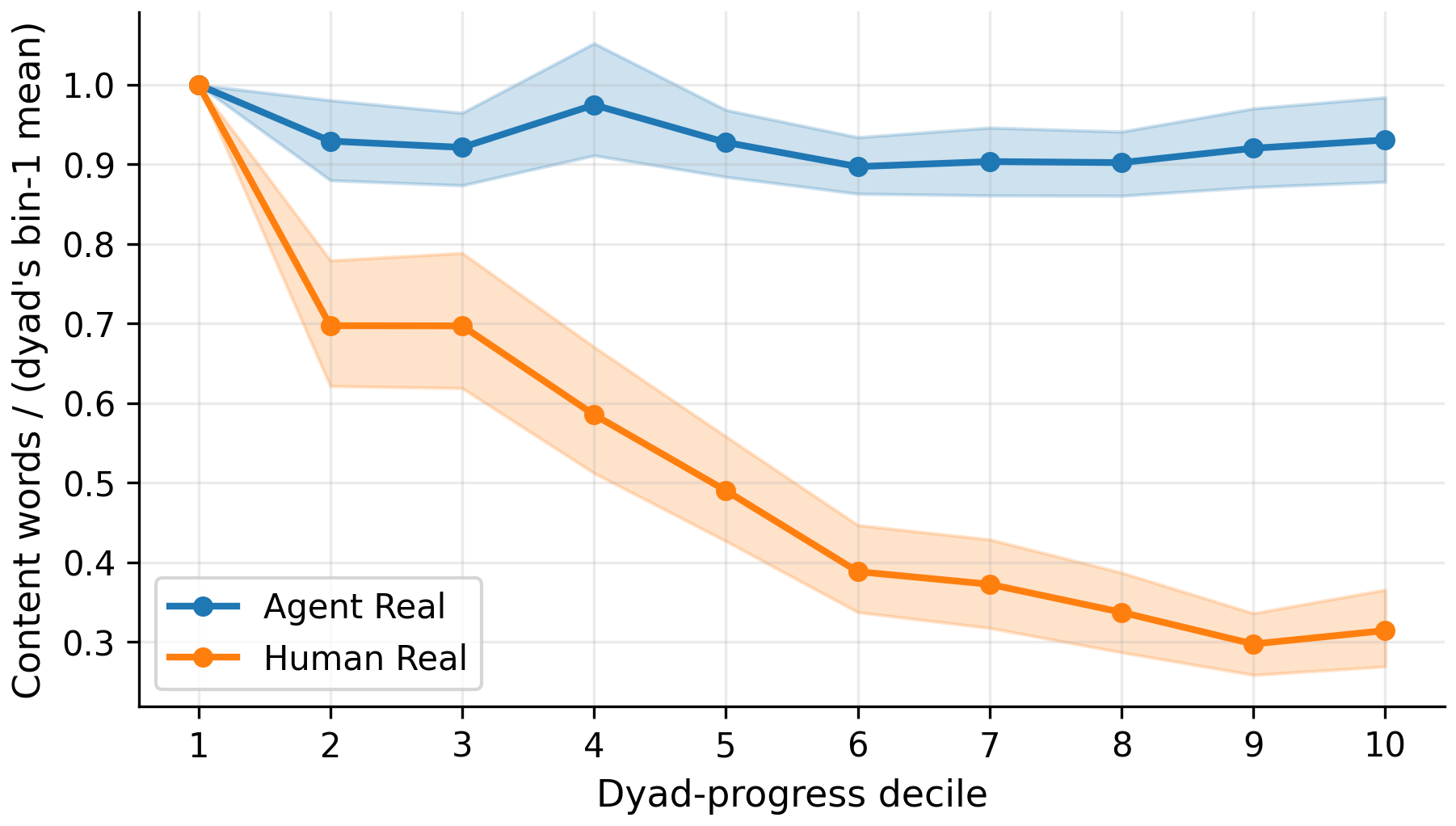}}
  \caption{Proportional content-word compression across dyad-progress deciles. Each curve shows mean content-word count per round normalised by that dyad's decile-1 mean, with 95\% bootstrap CI bands.}
  \label{fig:content-words}
\end{figure}

Humans compress proportionally across rounds; agents do not. By
mid-trajectory, human dyads produce roughly 50\% of their decile-1 content-word volume, dropping to $\sim$30\% by decile~9; agent dyads remain at $\sim$90\% of their starting level throughout
(Figure~\ref{fig:content-words}). A log-scale mixed-effects model recovers this as a significant group-by-round interaction ($\hat{\beta} = -0.303$, $p \ll 0.005$): the human log-slope is large and reliable ($\hat{\beta} = -0.372$), while the agent log-slope is near-zero and non-significant ($\hat{\beta} = -0.069$, $p = 0.20$). Humans thus compress proportionally to their current level, the signature of conceptual-pact formation, whereas agents start with descriptions already detailed enough for correct identification and never compress. This converges with the flat effort trajectory in Section~\ref{sect:task_comp_effort}: agents are verbose and rely on exhaustive description rather than on partner-specific labels.

\paragraph{Verbosity without an explicit efficiency pressure.}
Our task intentionally imposes no verbosity cost. The design mirrors the KTH human instructions, which set no brevity requirement and no explicit time limit: participants may communicate as much as they need, and the only temporal constraint \citet{shore2018kth} report is the 15-minute session length. Our goal was to test whether human-like adaptation emerges \emph{spontaneously} under instructions closely matched to the human experiment, not to elicit it through token budgets, brevity rewards, or instructions to reuse labels; such interventions address a different question, and are important future work. The absence of a verbosity cost plausibly explains why agent descriptions stay long, but not the absence of \emph{partner-specific} alignment: even holding descriptions verbose, an interaction-history-dependent account predicts greater alignment in real than pseudo dyads. Humans show that difference; agents do not (Section~\ref{subsec:micro_alignment}).

\subsection{Alignment Dynamics}
\label{subsec:micro_alignment}

\begin{figure}[tbp]
  \centering
  \includegraphics[width=\linewidth]{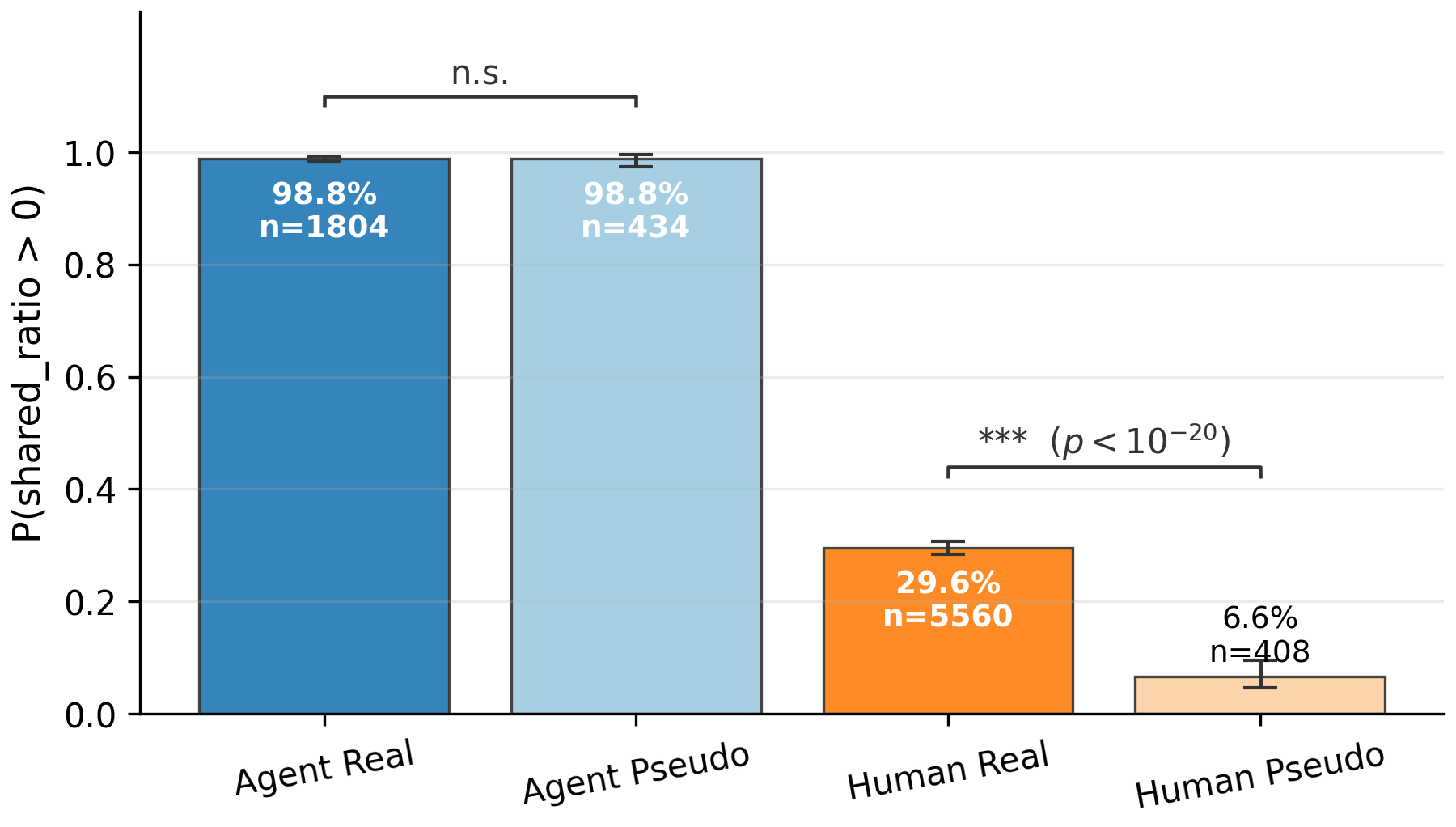}
  \caption{Turn-level ratio distributions for agent and human dyads.}
  \label{fig:shared-ratio}
\end{figure}

First, we investigate lexical alignment at the turn level. For each speaker-round, we compute the turn-level overlap ratio
\begin{equation}
\textsc{shared\_ratio} =
\frac{\textsc{shared turns}}{\textsc{total turns}} ,
\label{eq:shared}
\end{equation}
where shared turns are turns containing at least one lexical core
from the dyad's lexical-core inventory.
Figure~\ref{fig:shared-ratio} reports the probability that a
(speaker, round) cell contains at least one shared core.

Under the pseudo-dyad control, human and agent overlap diverge. In humans, real dyads are far more likely than pseudo dyads to produce a shared-lexical-core turn ($\hat{\beta}_{\text{logit}} = 1.92$, 95\% CI $[1.87, 1.97]$, OR $= 6.85$ $[6.51, 7.20]$): roughly one in five turns in a real human interaction contains a shared lexical core, whereas pseudo-paired human dialogues rarely do. In agents, overlap is near-ceiling in both conditions (real mean 0.979, pseudo mean 0.982; both medians 1.000), and the GLMM does not separate real from pseudo (OR $= 1.26$, 95\% CI $[0.99, 1.61]$). 
% Human lexical alignment thus depends on shared history; agent overlap persists when that history is removed.

Two factors explain the verbosity-driven agent ceiling. First, agents
draw on pretrained vocabulary heavily enough that the tangram shapes themselves, not partner-specific convergence, drive lexical-core activation. Second, agent turns are roughly eight times longer than human turns by token count, making it near-inevitable that at least one shared core appears in every utterance. Where humans develop shorthand labels (``the batman''), agents generate exhaustive geometric descriptions (``lime-green three-pronged crown sitting on a wide, low rectangle with a notch''), producing pervasive coverage by sheer volume. Human alignment is therefore cumulative and interaction-dependent; agent alignment is a byproduct of shared pretrained priors and descriptive verbosity, consistent with \citet{mayor2025llm} and \citet{shaikh2024grounding}.

% \emph{The distributions of pseudo and real agents do not differ, but the distributions of pseudo and human agents are significantly different.}

\paragraph{Reuse-intensity trajectories.}
Eq.~\ref{eq:shared} asks only \emph{whether} a turn contains any shared core. To quantify \emph{how much} shared material is produced over time, we count lexical-core occurrences in dyad $d$ at round $r$, $\text{Occ}(d, r)$, and rescale each round's count by that dyad's own busiest round:
\begin{equation}
\text{RelOcc}(d, r) =
\frac{\text{Occ}(d, r)}{\max_{r'} \text{Occ}(d, r')}.
\label{eq:relocc}
\end{equation}
$\text{RelOcc}(d, r) \in [0, 1]$ is therefore the dyad's reuse intensity at round $r$ as a fraction of its own peak: 1.0 marks the round at which the dyad reuses the most shared material, and all other rounds are scaled relative to it. Normalizing within each dyad removes the large baseline differences in absolute counts between humans and agents, so the shape of the trajectory, not its absolute height, becomes the object of comparison.

\begin{figure}[tbp]
    \centering
    \includegraphics[width=\linewidth]{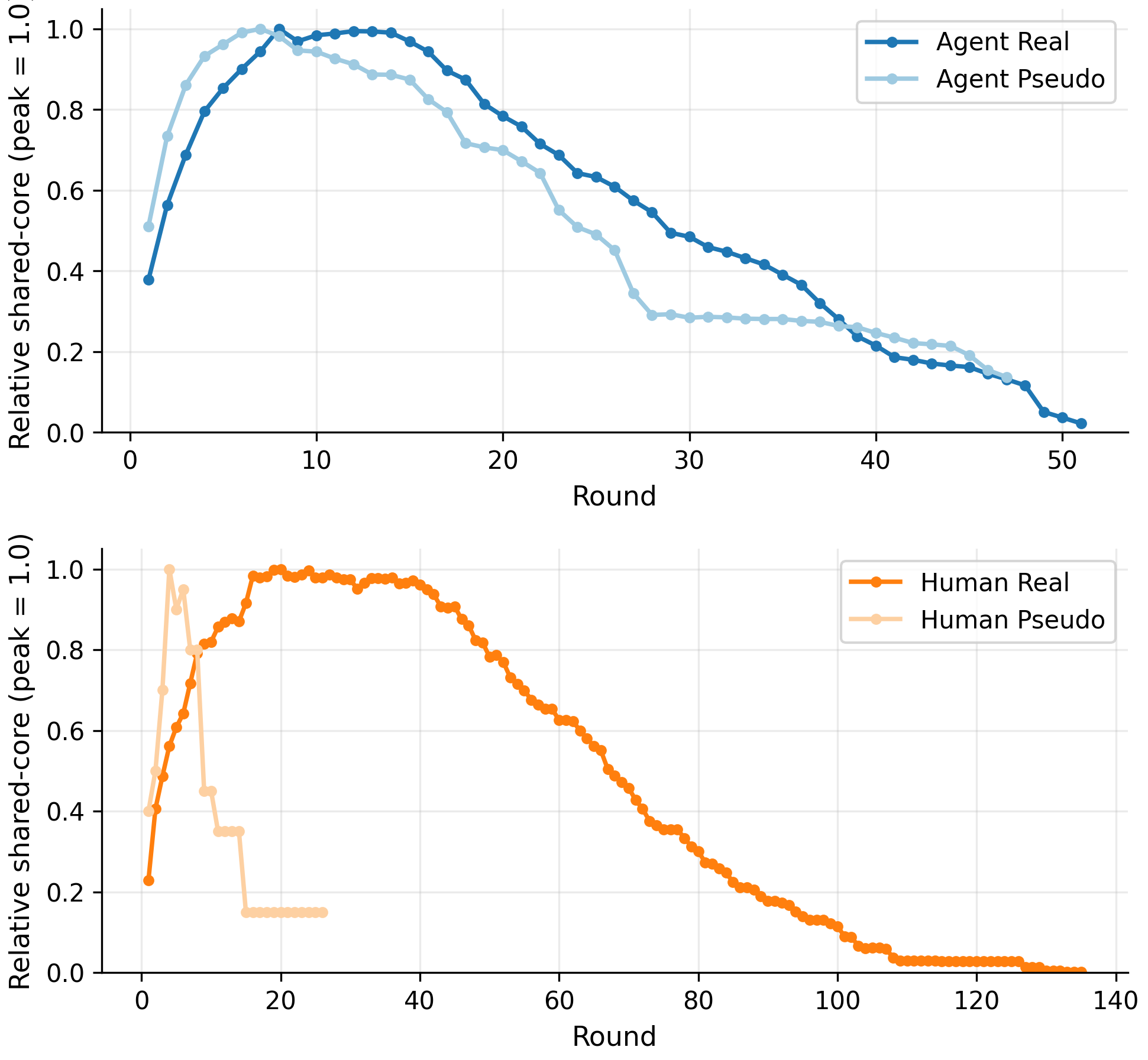}
    \caption{Reuse-intensity trajectories, peak-normalized within each dyad (Eq.~\ref{eq:relocc}).}
    \label{fig:perdyad}
\end{figure}

Both real datasets show an inverted-U (Figure~\ref{fig:perdyad}).
Agent real trajectories rise sharply from round~1, peak at rounds~7--10, and decay toward zero by round~50; agent pseudo follows a similar but lower trajectory, with only modest separation in the middle rounds. Human real trajectories rise gradually, peak at rounds~5--10, and sustain activity through round~80+ before tapering near round~130; human pseudo-trajectories peak sharply in the first few rounds and collapse by round~15. The large human-real/pseudo gap is the strongest corpus-level evidence that lexical-core reuse in humans depends on interactive grounding, while the weaker agent separation indicates that interaction history contributes only marginally beyond what pretrained priors already provide.

We argue that current MLLMs show global style alignment rather than the partner-specific, history-dependent grounding characteristic of human conceptual pacts. They produce consistent, vocabulary-appropriate descriptions yielding high overlap across interlocutors whether they share an interaction history or not.

Partner-specific grounding emerges only marginally, through structural depth and sustained trajectory intensity, rather than through the compact, flexibly renegotiable conceptual pacts of human dialogue.
These findings are consistent with \citet{zeng2026lvlmsACL}, who find that LVLMs fail to establish common ground in a factorial referential design; \citet{imai2025measuring}, who document lexical adaptation deficits despite adequate content alignment; and \citet{kwon2025evaluating}, who observe style matching but strategic divergence from humans. Our results extend this literature by showing that agents can achieve high task success and turn efficiency while still lacking the hallmarks of interaction-specific grounding, underscoring that raw performance metrics are insufficient for evaluating cooperative dialogue competence.

\section{Conclusion}
\label{sec:conclusion}
% \amber{I try to make it tighter}
Across three measures (task competence and effort, description length, and lexical reuse), MLLM agent dyads diverge from human dyads in how they achieve referential coordination: agents maintain fixed effort and near-ceiling lexical overlap, which are statistically indistinguishable between real and pseudo conditions, whereas humans reduce effort, compress descriptions, and develop partner-specific lexical-core reuse that grows over time. In this task, MLLMs thus achieve coordination without conceptual pacts.

The main implication is methodological. Task success and raw lexical overlap remain high even when partner history is removed. A shared-history control is therefore necessary when alignment in agent dialogue is used as evidence of grounding. 

Our pragmatically constrained pseudo-dyad algorithm, three-layer analytic protocol, and lexical-core extraction pipeline are released as a reusable framework. The pseudo-dyad baseline is model-agnostic, i.e., the dissociation reported here is one instantiation, and the same diagnostics can be reapplied as new models and training regimes become available.

\section*{Limitations}

\paragraph{Model generalizability.}
Our agent experiments use GPT-5 because its Responses API provides infrastructure-level lossless multimodal interaction history across rounds, a non-trivial requirement for studying lexical convergence. The pseudo-dyad methodology and analysis pipeline are model-agnostic, and we release them to facilitate replication.

Related findings suggest that this descriptive-verbosity pattern may not be specific to GPT-5. \citet{hua2024talk} report similar behavior across multiple MLLMs.\citet{vaduguru2026success} show that convention formation requires explicit cost-sensitive training objectives that are absent from standard instruction tuning. Together, these findings suggest that the pattern may result from the success-only optimisation regime shared by current instruction-tuned MLLMs. Replication on open-weight models is a priority for future work.

\paragraph{Shorter agent co-reference chains.}
On average, human dyads mention a given tangram 3.93 times per session (median 4, maximum 15). In contrast, agent dyads average just 1.72 mentions because their sessions halt at the first label violation. This creates an asymmetry: agents experience a narrower co-reference range and receive fewer opportunities to compress their descriptions. This disparity does not produce a false dissociation, as the difference is visible at the very first repeated reference: humans compress by the second mention, while agents show limited reduction (Appendix~\ref{app:coref}). Still, observing longer agent sessions would provide a stronger test for the later, more conventionalized parts of the interaction.
\paragraph{A single shared visual environment.}
We expose all 45 agent dyads to the same 80-view sequence, taken from a single human KTH session (session~26), but with staggered starting points. This strategy holds the visual context constant, which is necessary for exact-piece pseudo-dyad matching. The trade-off is a loss of diversity relative to the human corpus. Furthermore, because sessions stop at different times, image coverage across the agent dyads is uneven. Testing these agents on sequences from various human sessions would help confirm that our observed patterns are not just an artefact of this single visual setup.
\paragraph{No human--agent condition.}
We compare human--human and agent--agent dyads but do not include human--agent interaction. The factorial design of \citet{zeng2026lvlmsACL} demonstrates the value of such conditions; testing our metrics in human--AI settings is a direct next step.

\paragraph{Modality asymmetry.}
The KTH corpus consists of spoken dialogue; our MLLM dyads interact via text. However, cross-round compression, partner-specific conceptual pacts, and reduction have been robustly replicated in large-scale text-based tangram games \citep{hawkins2020characterizing}, so conceptual-pact formation is not an artefact of the spoken modality. The descriptive verbosity we observe cannot be explained away as a mismatch between spoken human and text-based agent baselines.
\paragraph{Scope of referential games.}
By using a game-like structure, referential games successfully generate spontaneous dialogue and model how humans refer to things in everyday life. Even so, they remain an artificial setup. As a result, researchers should be careful when applying results from these games, including the findings in this study, to fully open-ended communication.

\section*{Ethical Considerations}

In alignment with its intended purpose, we utilize the KTH Tangrams human corpus \citep{shore2018kth} exclusively for scientific benchmarking and research. This established public dataset consists solely of task-oriented conversations about abstract shapes and contains no personally identifying information (PII). Similarly, the novel MLLM agent dataset we introduce is restricted to research use, adhering to the original data access terms. A manual review of a corpus sample confirmed the complete absence of PII and offensive material. While we anticipate no direct risks of misuse, we note a potential interpretive risk: findings like ours can easily be exaggerated into broad claims about machine comprehension. To mitigate this, we have deliberately constrained our conclusions (Sections~\ref{sec:method-pseudo} and~\ref{subsec:micro_alignment}) to the specific behaviors exhibited within the parameters of this task.

% \section*{Acknowledgments}

\bibliography{custom}

\appendix

\section{GPT-5 Parameter Sweep}
\label{app:config}

\begin{figure}[tbp]
  \centering
  \subfloat[Exact Match]{\includegraphics[width=0.48\linewidth]{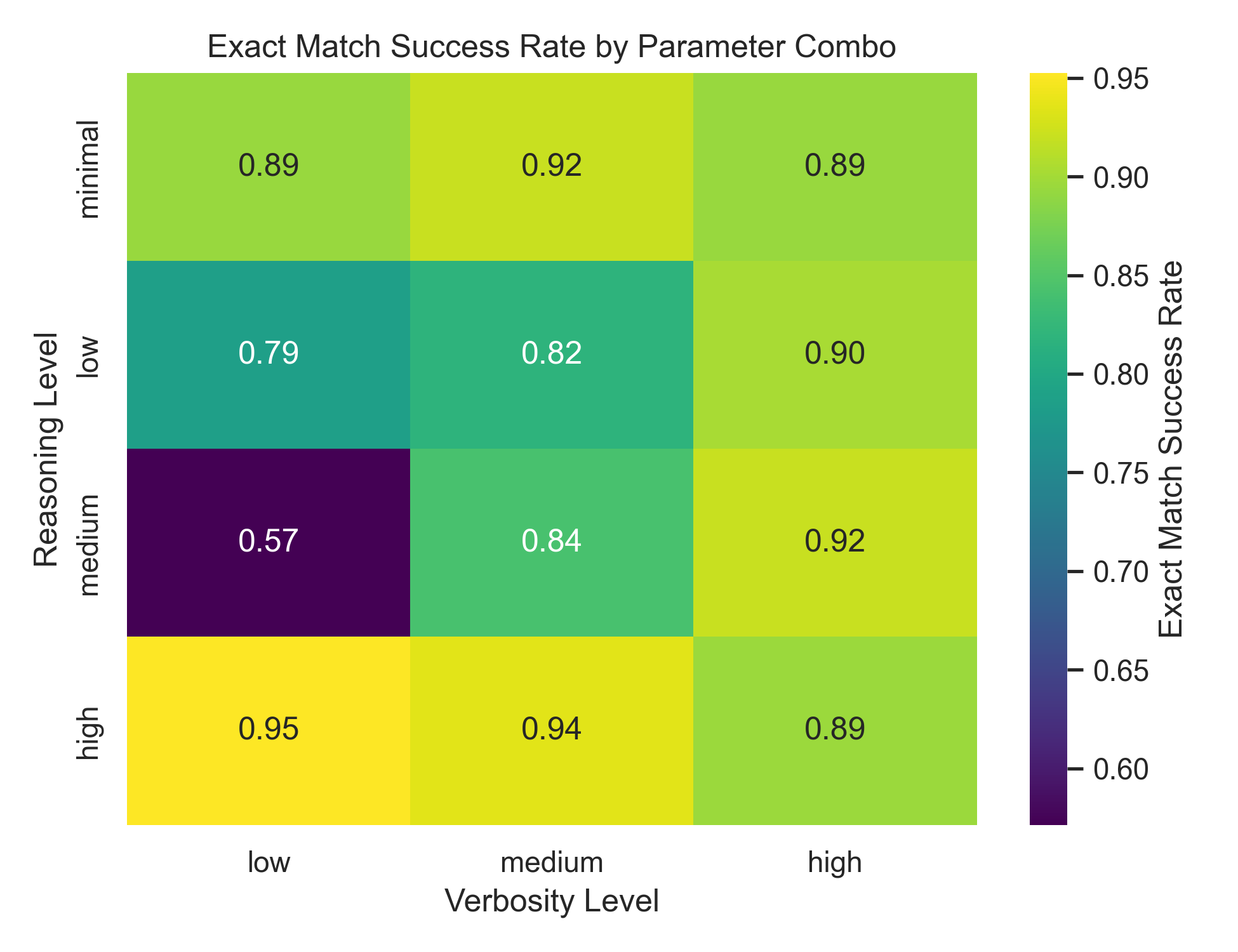}}\hfill
  \subfloat[Label Violations]{\includegraphics[width=0.48\linewidth]{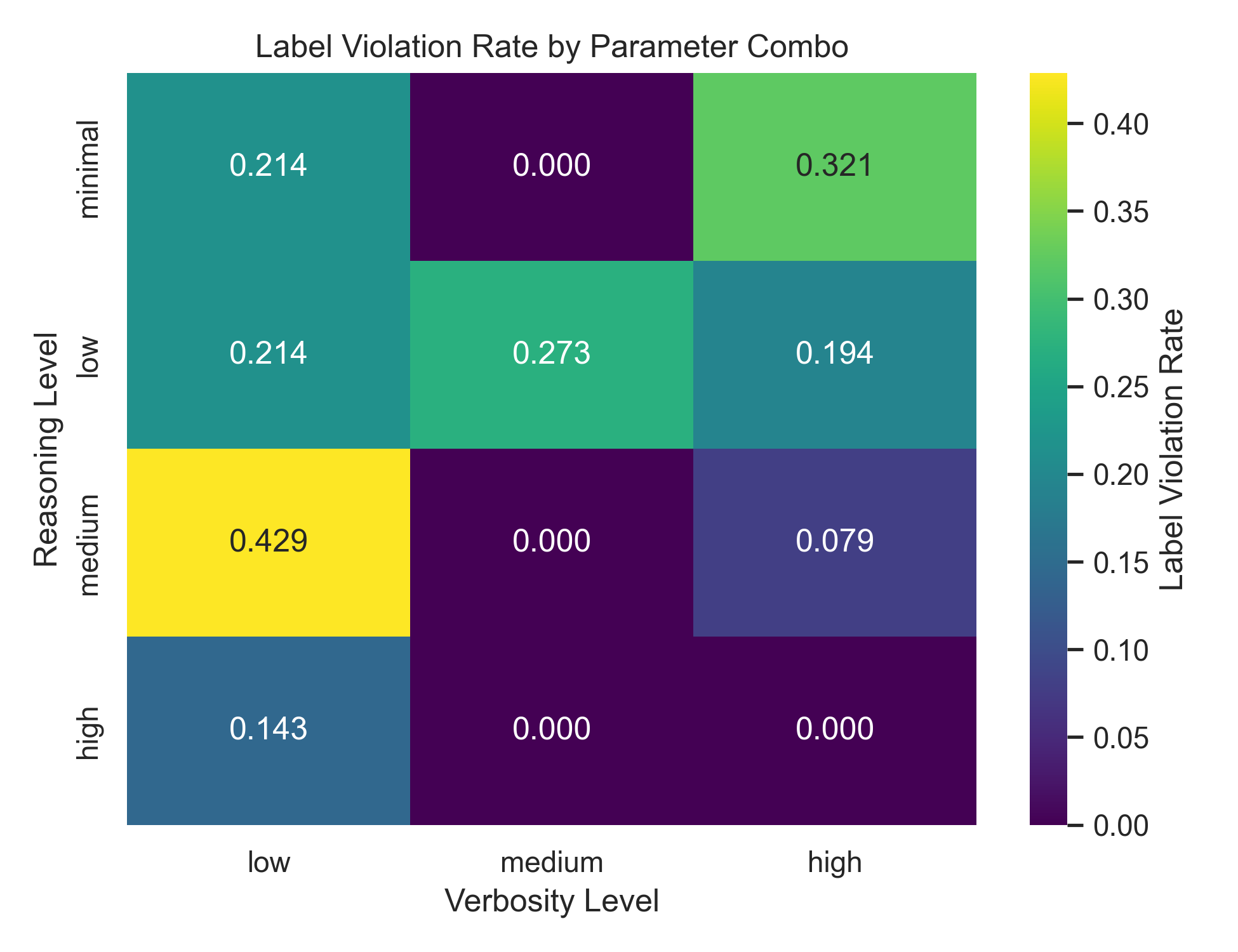}}\\[4pt]
  \subfloat[Turn Economy]{\includegraphics[width=0.48\linewidth]{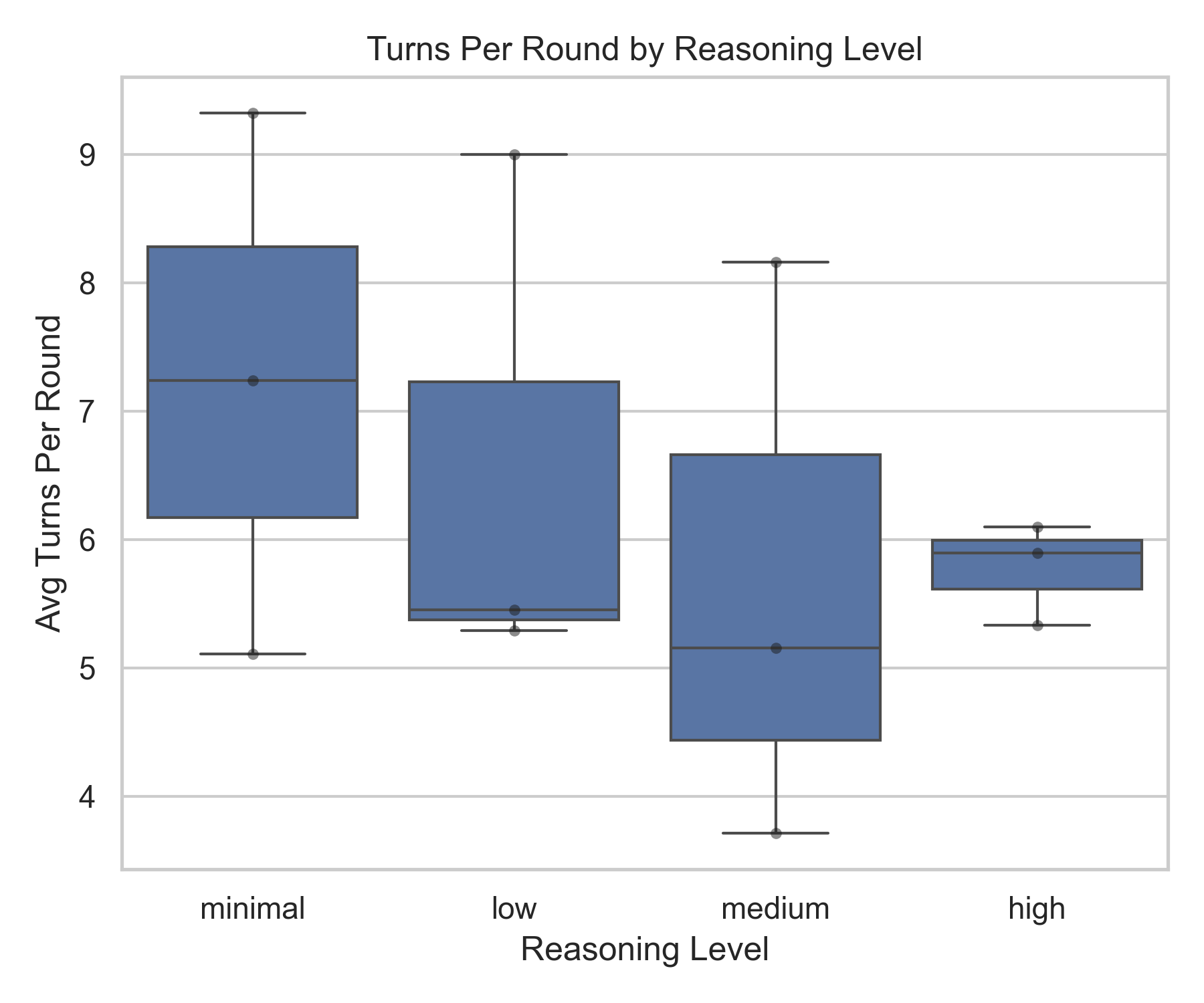}}\hfill
  \subfloat[Failure Types]{\includegraphics[width=0.48\linewidth]{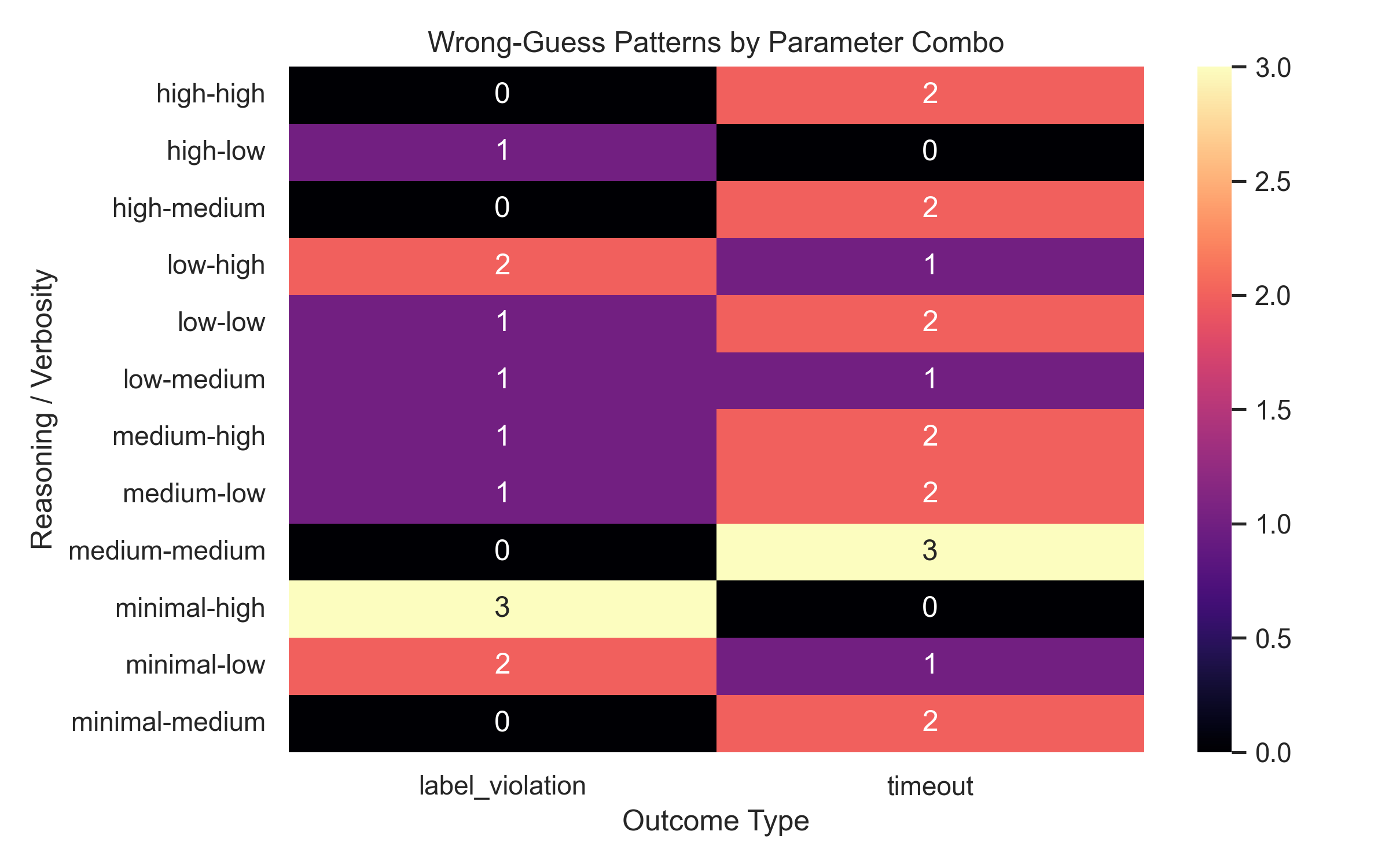}}
  \caption{Configuration performance across the reasoning $\times$ verbosity parameter sweep.}
  \label{fig:config}
\end{figure}

We swept GPT-5 (\texttt{gpt-5-2025-08-07}) over reasoning effort $\in \{\text{minimal, low, medium, high}\}$ and verbosity $\in \{\text{low, medium, high}\}$, yielding 12 configurations with 3 sessions each (36 runs total). Figure~\ref{fig:config} reports exact-match accuracy, label violations, turn economy, and failure-type distributions across the sweep. The adopted configuration (high reasoning, medium verbosity) achieves zero label violations and 94\% exact-match accuracy.

\section{View Pairing and Auto-Labelling Pipeline}
\label{app:pipeline}

This appendix expands on the implementation details for resolving the materials-preparation challenges introduced in Section~\ref{sec:method-adapt}. Because the KTH dataset stores informant and guesser perspectives with inconsistent filenames and screenshot sizes, automated labeling is non-trivial. To overcome this, we developed an automatic pairing and labelling pipeline. The pipeline (i) matches informant and guesser views using weighted colour histograms, template matching, and ORB keypoints (Oriented FAST and Rotated BRIEF; \citealp{rublee2011orb}), and (ii) assigns randomised two-letter labels to the centroid of each shape using the Hungarian algorithm \citep{kuhn1955hungarian}. After manual checks, the pipeline resulted in 80 informant--guesser view pairs. Because these 80 pairs are consecutive screenshots from a single human session (session~26), they inherently preserve that session's specific board dynamics, where only the successfully identified piece shifts position.

\section{Prompts for Guiding Agents in the Tangram Task}
\label{app:prompts}

This appendix reproduces the full set of natural-language prompts used
to instantiate the agent dyads. All prompts were held constant across
all 45 dyads and all 905 rounds. We distinguish (i)~the shared system prompt, (ii)~functional-role definitions for the informant and guesser,
(iii)~the judge prompt used for outcome classification, and (iv)~the system messages issued by the controller. Outcomes are scored by exact match on the two-letter label; embedding-similarity and LLM-judge verdicts were logged alongside but contributed no points and are not used in any reported result.

\subsection{System prompt}
\label{app:prompts-system}

The same system prompt is shown to both agents at session start. 

\begin{quote}\small
\textbf{[Tangram]} Thank you for participating in the study! You will be playing an online game with another player. You both will be presented
with a board of tangrams.

\medskip
\textbf{[Play]} Each player has one of two roles: either the
``informant''---the player who can see the target piece, which is highlighted with a dotted box---or the ``guesser''---the player who must guess the target piece. You will receive instructions about your current role. After guessing, both the informant and guesser will receive feedback from the computer regarding the correctness of the
guess and their current accumulated score.

\medskip
\textbf{[Goal]} The goal of the game is to score as many points as possible by correctly guessing the target piece. By guessing the correct piece, you gain one point. If you guess the wrong piece, you lose two points but are allowed to try again. The two players are allowed to communicate as much as they want before each guess in order to help the guesser identify the correct piece. Never mention the labels in the picture unless the computer asks. 
\end{quote}

For different roles, the role setting is attached after the aforementioned shared system prompt.

\paragraph{Informant.}
You are the INFORMANT in a tangram guessing game. Look at the image and describe the target piece marked with a dotted box around it. This is the piece which the other player must guess. Your goal is to help the guesser identify it. Provide any details or answer any questions that the guesser may ask about the piece they are trying to guess. Never mention the labels in the picture unless asked by the computer! If the guesser's guess is incorrect, provide additional details about the marked piece to help the guesser identify it correctly. You will be notified of your accumulated score and the correctness of the guesser's guess by the computer after each guess. 
\paragraph{Guesser.}
You are the GUESSER. Based on the informant's description, conversation history, the informant's answers to your questions, and the image you see, ask clarification questions before you make a guess or guess the piece directly.

\begin{enumerate}
\item Please never mention the labels in the picture unless asked by the computer!
\item If you think you know which piece it is, you must say "I know the answer!" (include this exact phrase, do not mention the labels) and specify why you think it's that one.
\item If you're not sure, ask specific questions to help with your identification.
\item You can ask as many questions as you need to correctly guess the piece.
\item You will be notified of your accumulated score and the correctness of your guess by the computer after each guess.
\end{enumerate}

\subsection{Functional-role prompts}
\label{app:prompts-roles}

Each turn is generated by issuing one of the following role-bound action prompts to the relevant agent. The prompts deliberately avoid explicit instructions about turn length, label reuse, or convention formation; the only hard constraint is the non-leaked instruction about the two-letter tags.

\paragraph{Informant.}

\begin{itemize}
\item \texttt{describe}: ``ROLE: Informant. Please follow the system prompt's role-setting instructions. Never mention the labels in the picture unless asked by the computer!''
\item \texttt{provide\_label}: ``ROLE: Informant. What is the label for
the piece you identified? Respond with just the label.''
\item \texttt{provide\_feedback}: ``ROLE: Informant. The guesser
guessed the incorrect piece. The current score is now updated. Provide additional details about the marked piece to help the guesser identify it correctly. Never mention the labels in the picture.''
\item \texttt{natural\_response}: (no instruction, used to elicit free-form continuation when neither a label nor feedback is required.)
\end{itemize}

\paragraph{Guesser.}

\begin{itemize}
\item \texttt{make\_guess}: ``ROLE: Guesser. Please follow the system prompt's role-setting instructions. Never mention the labels in the picture unless asked by the computer!''
\item \texttt{provide\_label}: ``ROLE: Guesser. What is the label for
the piece you identified? Respond with just the label.''
\end{itemize}

\subsection{Computer (system) messages}
\label{app:prompts-system-msgs}

The computer\footnote{ To avoid confusion, we referred to the computer system message as the controller in the previous discussion.} issues the following deterministic messages at round boundaries; these are not generated by an LLM.

\begin{itemize}
\item \texttt{correct\_guess}: ``SYSTEM. The guesser guessed the correct piece. The current score is \texttt{\{score\}}. You are going to start a new round.''
\item \texttt{incorrect\_guess}: ``SYSTEM. The guesser guessed the
incorrect piece. The current score is \texttt{\{score\}}. The informant is going to provide extra information.''
\end{itemize}

\section{Lexical-Core Construction}
\label{app:cores}

This appendix gives the filtering and clustering details summarised in Section~\ref{sec:method-cores}. After detecting repeated expressions (i.e., words), we refine and process them as follows. First, apply a content-based filter that removes: (a) highly frequent words that are too frequent and automatized to serve as informative referential content; (b) function words; and (c) shared surface forms that contain no content word or have a content-word ratio below 0.2. Second, the remaining surface forms are clustered using a union-find algorithm based on their content-lemma sets. Two surface forms are merged when they share at least one content lemma and their Jaccard overlap is at least 0.6:
\begin{equation}
\textsc{overlap}(S_1, S_2) =
\frac{|L_1 \cap L_2|}{|L_1 \cup L_2|} \geq 0.6 ,
\end{equation}
where $L_1$ and $L_2$ denote the content-lemma sets of surface forms $S_1$ and $S_2$, respectively. Each resulting cluster is assigned a label based on the alphabetized intersection of its lemmas; when this intersection is empty, the most frequent surface form in the cluster is used instead. These lexical cores constitute the unit of analysis for all subsequent alignment metrics.

It is important to emphasize that a lexical core serves purely as an operational approximation for a conceptual pact, rather than representing the pact itself. While it successfully measures the surface-level linguistic convergence that typically results from such an agreement, it cannot guarantee that both speakers have genuinely reached a mutual understanding of the referent.
\section{Co-reference Chain Lengths}
\label{app:coref}

Because a \emph{round} constitutes a single target-identification event rather than a complete cycle through the board (Section~\ref{sec:method-terms}), the number of times a piece becomes the target within a session is an empirical property of the interaction rather than a predetermined design parameter. We report this distribution for both populations below.

At piece level, a \emph{dyad--piece chain} is the ordered sequence of all rounds within one dyad's session in which the same within-session piece was the target; an \emph{observed} chain contains at least one such round. The human corpus contains 836 observed dyad--piece chains across 42 sessions and 3{,}288 rounds, with an average of 3.93 target rounds per chain (median~4, maximum~15). When chains are instead keyed by shape outline, the 20 pieces on each board collapse to 17 outlines, with three outlines instantiated by two pieces differing in size and/or colour. The resulting 710 observed dyad--outline chains contain an average of 4.63 target rounds (median~4, maximum~21). These chain lengths align with \citet{shore2018kth}, whose analyses cover coreference orders 2--6 and document within-dialogue referent chains spanning 4--15 mentions (their Figures~4--6 and Table~7). Agent chains are shorter because the controller halts an interaction at the first label violation, which terminated each of the 45 agent sessions. The agent corpus contains 525 observed dyad--piece chains, with an average of 1.72 target rounds per chain (median~2, maximum~5).

For the first--second mention comparison, we define the $k$-th mention of a piece as the $k$-th logged target-identification round within that dyad's session in which the same within-session piece was the target. For this comparison, we restrict the sample to dyad--piece chains containing at least two target rounds; thus, $n$ denotes the number of matched chains contributing one first-mention and one second-mention observation each. We use the informant's round-level conversational word count to approximate description length, excluding system, label, and guesser turns. A transcribed guesser turn is not required for a round to enter the primary analysis. Under this definition, the human mean decreases from 28.97 to 15.26 words, a reduction of $47.3\%$ ($n=751$), whereas the corresponding agent mean decreases from 71.70 to 69.09 words, a reduction of only $3.6\%$ ($n=263$).\footnote{As a sensitivity analysis, requiring at least one transcribed conversational turn from both roles in both the original first- and second-mention rounds produces similar estimates: a $45.5\%$ reduction for humans ($n=533$) and a $2.9\%$ reduction for agents ($n=261$).}

These findings have two implications. First, agent trajectories cover a narrower range of coreference orders, which we acknowledge as a limitation. Second, the human--agent adaptation contrast is already evident at the first repeated reference and therefore cannot be attributed solely to the longer tail of the human chain-length distribution.

\section{Pseudo-Dyad Scoring and Tier Cascade}
\label{app:pseudo-scoring}

\paragraph{Constraints and score.}
Each candidate pair must use the same referent key, a shape outline for humans and an exact piece for agents, and satisfy the tier-specific round-position bound; neither constraint is relaxed. For each source round, Tiers~1--2 use three fallback stages: the first enforces both the total-turn bound and success parity, the second removes the total-turn bound, and the third also relaxes success parity. Tier~3 does not enforce success parity and therefore uses only the first two stages.

Writing the round and total-turn bounds as $(\delta_r,\delta_t)$, their values are $(1,3)$, $(2,5)$, and $(2,8)$ at Tiers~1--3. Thus, $|\Delta r|\leq\delta_r$ always, whereas $|\Delta t|\leq\delta_t$ only before the total-turn bound is relaxed. For agent matches, each role-specific turn-count difference must also be at most one. Human role-specific differences greater than two are permitted but logged as high imbalances.

For an admissible pair $(r_a,r_b)$, the implementation computes
\begin{equation}
\label{eq:score}
\begin{aligned}
s(r_a,r_b)={}&w_r\lvert\Delta r\rvert
+w_t\lvert\Delta t\rvert\\
&+\lvert\Delta t_{\mathrm{inf}}\rvert
+\lvert\Delta t_{\mathrm{gue}}\rvert\\
&+w_sM_{ab}.
\end{aligned}
\end{equation}
Here, $\Delta r$ is the round-index difference. The term $\Delta t$ is the difference in logged round-level total-turn counts, including system or label turns where present, whereas $\Delta t_{\mathrm{inf}}$ and $\Delta t_{\mathrm{gue}}$ compare retained conversational turns for the two roles. The indicator $M_{ab}$ equals one only when both source outcomes are observed and equal; it is zero when they differ or either outcome is missing. The weights $(w_r,w_t,w_s)$ are $(1.5,1.0,1.5)$ at Tier~1, $(1.0,1.0,1.0)$ at Tier~2, and $(0.5,0.5,0.0)$ at Tier~3.

\paragraph{Matching algorithm.}
For a fixed pair of source dyads, rounds from dyad $A$ are processed in order. Each source round independently traverses the fallback stages, and the first stage containing an admissible unused round from dyad $B$ is selected. Within that stage, the candidate with the smallest $s(r_a,r_b)$ is used and then removed from further consideration. Different rounds from the same dyad pair may therefore resolve to different stages.

The outcome term cannot affect this within-stage minimisation. In the success-constrained stages, every admissible candidate has $M_{ab}=1$. When a source round reaches the success-relaxed stage, no parity-matched candidate remains, so every available candidate has $M_{ab}=0$. Tier~3 instead sets $w_s=0$. The success-relaxed stage was used for 12 of 238 agent pseudo-rounds ($5.0\%$) and 1 of 204 human pseudo-rounds ($0.5\%$).

Candidates producing fewer than the tier minimum are discarded. The human minima are $(5,3,2)$ matched rounds at Tiers~1--3, and the agent minima are $(3,2,2)$. Among the remaining candidates, the algorithm selects the partner maximising the unnormalised aggregate $S(A,B)=\sum_{(r_a,r_b)}s(r_a,r_b)$. Thus, the per-round score is minimised locally, but its dyad-level sum is maximised. The aggregate is not a globally minimised distance and can favour more matches, larger admissible deviations, or both. Although the outcome term is inert
during round-level selection, it contributes at the dyad level: holding the other components fixed, each additional known-equal outcome increases the maximised aggregate by $w_s$. Dyads are processed longest-first, and each source dyad is paired at most once.

\paragraph{Matching diagnostics.}
Only rounds with valid role identities and at least one retained conversational turn from each role enter the pseudo-dyad source pool. Consequently, human rounds in which the guesser selected without speaking are excluded from pseudo-dyad construction. These rounds remain
part of the real human corpus. Specifically, the loader retains 2{,}272 of 3{,}288 human rounds; each of the excluded 1{,}016 rounds contains an informant turn but no logged guesser identity or guesser turn. The agent loader retains 899 of 905 rounds; the other six end after an informant label violation and contain no guesser turn.

Of the 238 agent pairs, 195 (82\%) match exactly on round index, 42 (18\%) differ by one, and one differs by two. Of the 204 human pairs, 85 (42\%) match exactly, 118 (58\%) differ by one, and one differs by two. Both source rounds have observed successful outcomes in 226 agent pairs (95.0\%) and 203 human pairs (99.5\%).

% The conversation-length comparison in Section~\ref{sec:method-pseudo} first calculates turns per round within each dyad and then averages those dyad-level rates; it is not derived from the pooled means in Table~\ref{tab:dataset}. Using the unrounded dyad-level means, agent pseudo-dyads have 14.7\% fewer turns per round
% than real agent dyads, whereas human pseudo-dyads have 15.0\% more.
The conversation-length comparison in Section~\ref{sec:method-pseudo} first calculates turns per round within each dyad and then averages those dyad-level rates, with the real-side baseline computed over the eligible rounds defined above. It is therefore not derivable from the pooled, all-round means in Table~\ref{tab:dataset}.

\section{Success-Only Corpus Statistics}
\label{app:success_only}
Table~\ref{tab:success_only} presents success-only statistics. Despite completing fewer rounds per dyad (19.11 vs.\ 78.12), agent dyads produced considerably more shared surface forms per dyad (104.02 vs.\ 16.74) and more word tokens overall (103,079 vs.\ 68,010), reflecting LLM verbosity: agent utterances averaged approximately 59~tokens per turn compared to 7~tokens for humans.

\begin{table}[t]
\centering
\small
\begin{tabular}{lrr}
\toprule
\textbf{Metric} & \textbf{Human Real} & \textbf{Agent Real} \\
\midrule
Dyads & 42 & 45 \\
Total rounds & 3,281 & 860 \\
Total turns & 10,228 & 1,742 \\
Mean rounds/dyad & 78.12 & 19.11 \\
Mean turns/round & 3.12 & 2.03 \\
Mean turns/dyad & 243.52 & 38.71 \\
Mean turns-to-success & 2.56 & 2.03 \\
Word tokens & 68,010 & 103,079 \\
Shared surface forms & 703 & 4,681 \\
Mean surface forms/dyad & 16.74 & 104.02 \\
Lexical cores & 503 & 2,888 \\
Mean cores/dyad & 11.98 & 64.18 \\
\bottomrule
\end{tabular}
\caption{Success-only rounds.}
\label{tab:success_only}
\end{table}

\section{Turns-to-Success Including Failed Rounds}
\label{app:tso_failure}

\begin{figure}[tbp]
  \centering
  \subfloat[All Rounds]{\includegraphics[width=\linewidth]{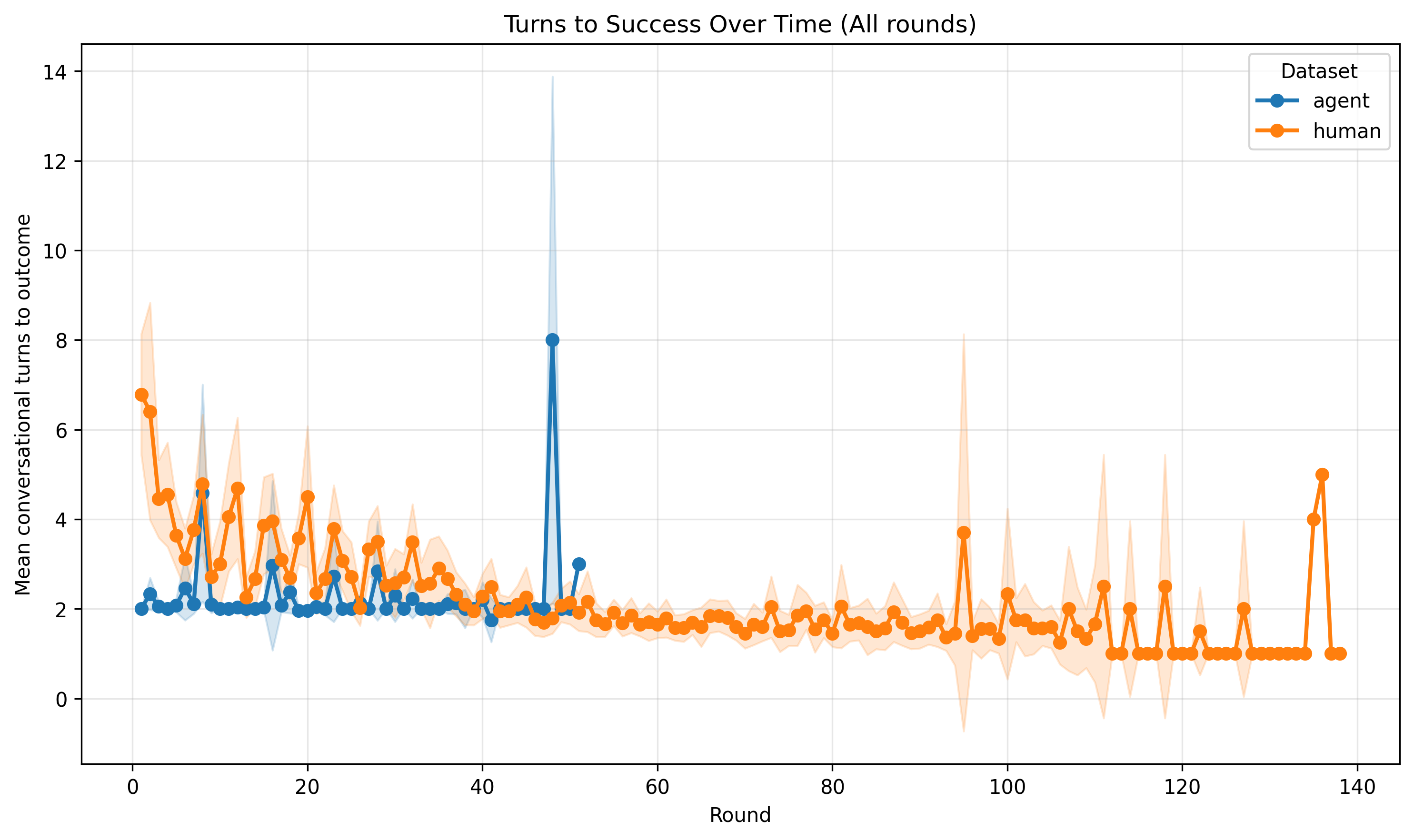}}\hfill
  \subfloat[First 20 Rounds]{\includegraphics[width=\linewidth]{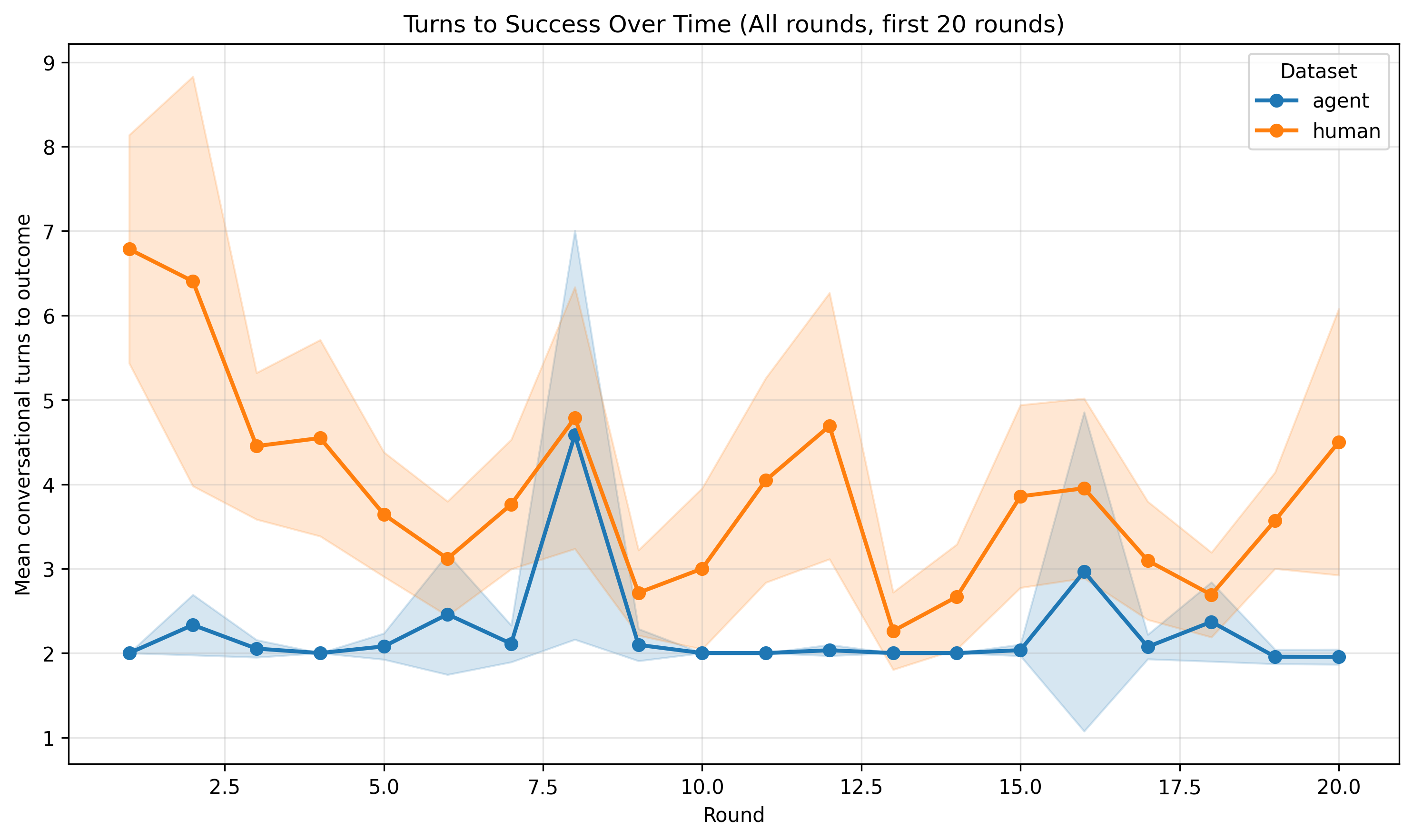}}
  \caption{Turns to Success Over Time (Failure Included).}
  \label{fig:TSO_f_all_20}
\end{figure}

Figure~\ref{fig:TSO_f_all_20} shows turns-to-success trajectories with failed rounds included. Spikes in this view correspond to failed rounds with extended back-and-forth.
% , complementing the success-only view in Figure~\ref{fig:TSO_all_20}. 

In the all-rounds panels, occasional spikes correspond to failed rounds involving extended back-and-forth before timing out or producing an incorrect guess. These spikes are absent from the success-only view, confirming that failed rounds tend to be substantially longer than successful ones.

\section{Aggregate Task Competence Plots}
\label{app:aggregate_plots}

\begin{figure}[tbp]
  \centering
  \includegraphics[width=\linewidth]{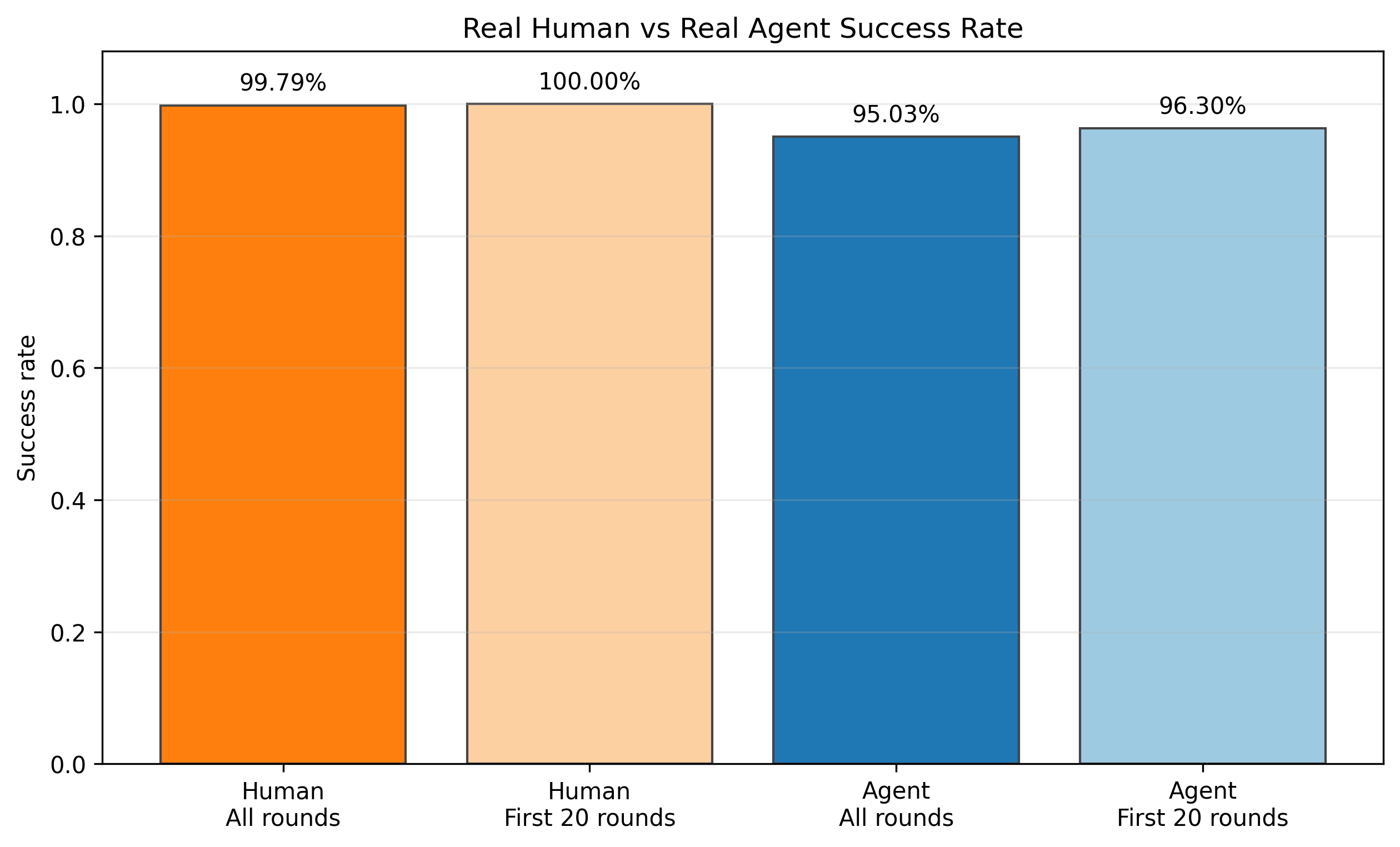}
  \caption{Success Rate.}
  \label{fig:sr}
\end{figure}

\begin{figure}[tbp]
  \centering
  \includegraphics[width=\linewidth]{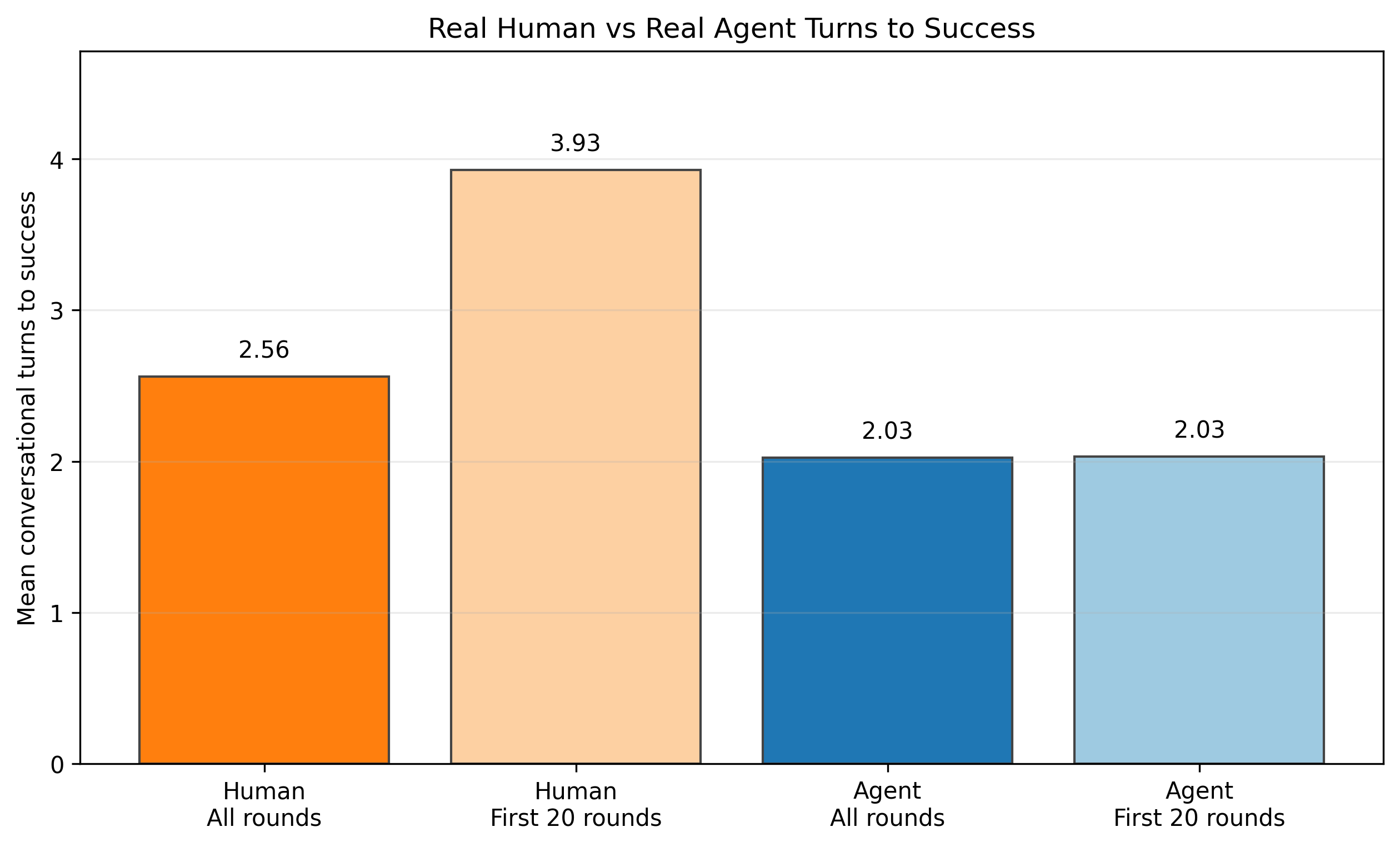}
  \caption{Turns to Success (aggregate).}
  \label{fig:ts}
\end{figure}

Figures~\ref{fig:sr} and~\ref{fig:ts} present the aggregate success rate and turns-to-success summaries discussed in Section~\ref{sec:results}.

\section{Total Word Trajectories}
\label{app:total_words}

\begin{figure}[tbp]
  \centering
  \subfloat[All Rounds]{\includegraphics[width=\linewidth]{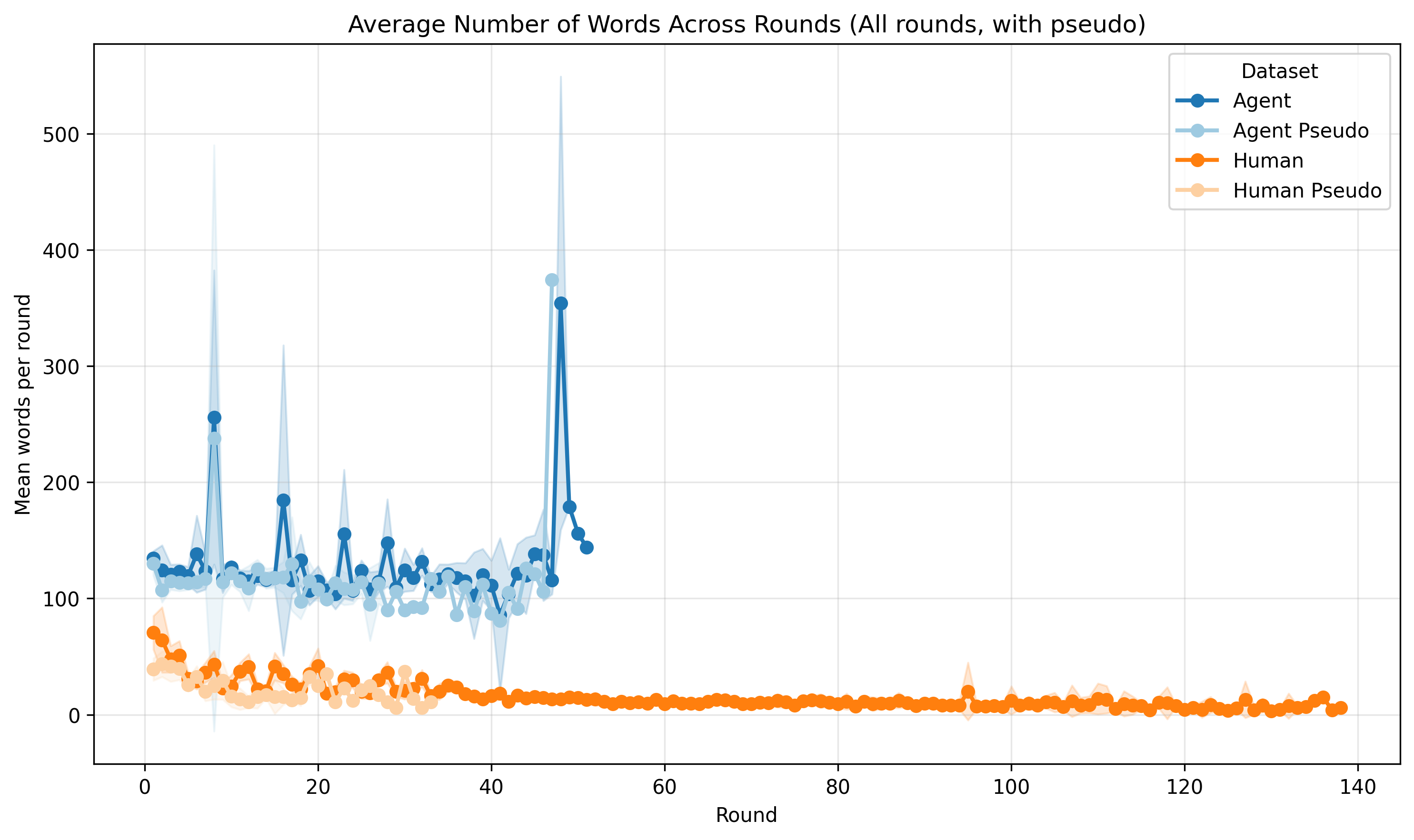}}\hfill
  \subfloat[Success Only]{\includegraphics[width=\linewidth]{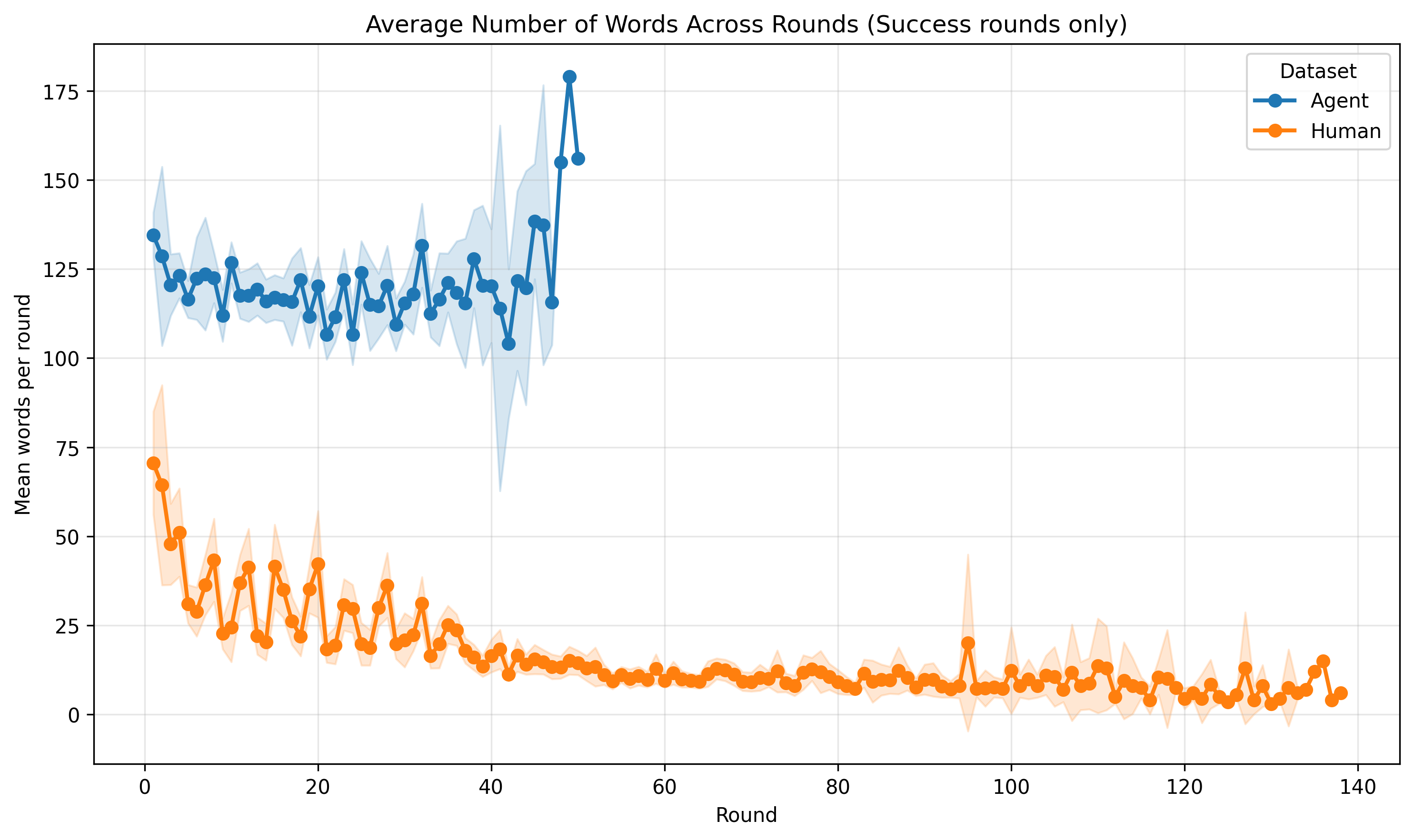}}
  \caption{Average Number of Words Across Rounds.}
  \label{fig:W_all_suc}
\end{figure}

Figure~\ref{fig:W_all_suc} shows total word counts across rounds, complementing the content-word trajectories in Figure~\ref{fig:content-words}. The total-word pattern mirrors the content-word pattern.

\section{Per-Dyad Shared-Turn Ratio over Rounds}
\label{app:perdyad}
\begin{figure}[tbp]
    \centering
    % Include the ESAM Plot 6B panels here
    \includegraphics[width=\linewidth]{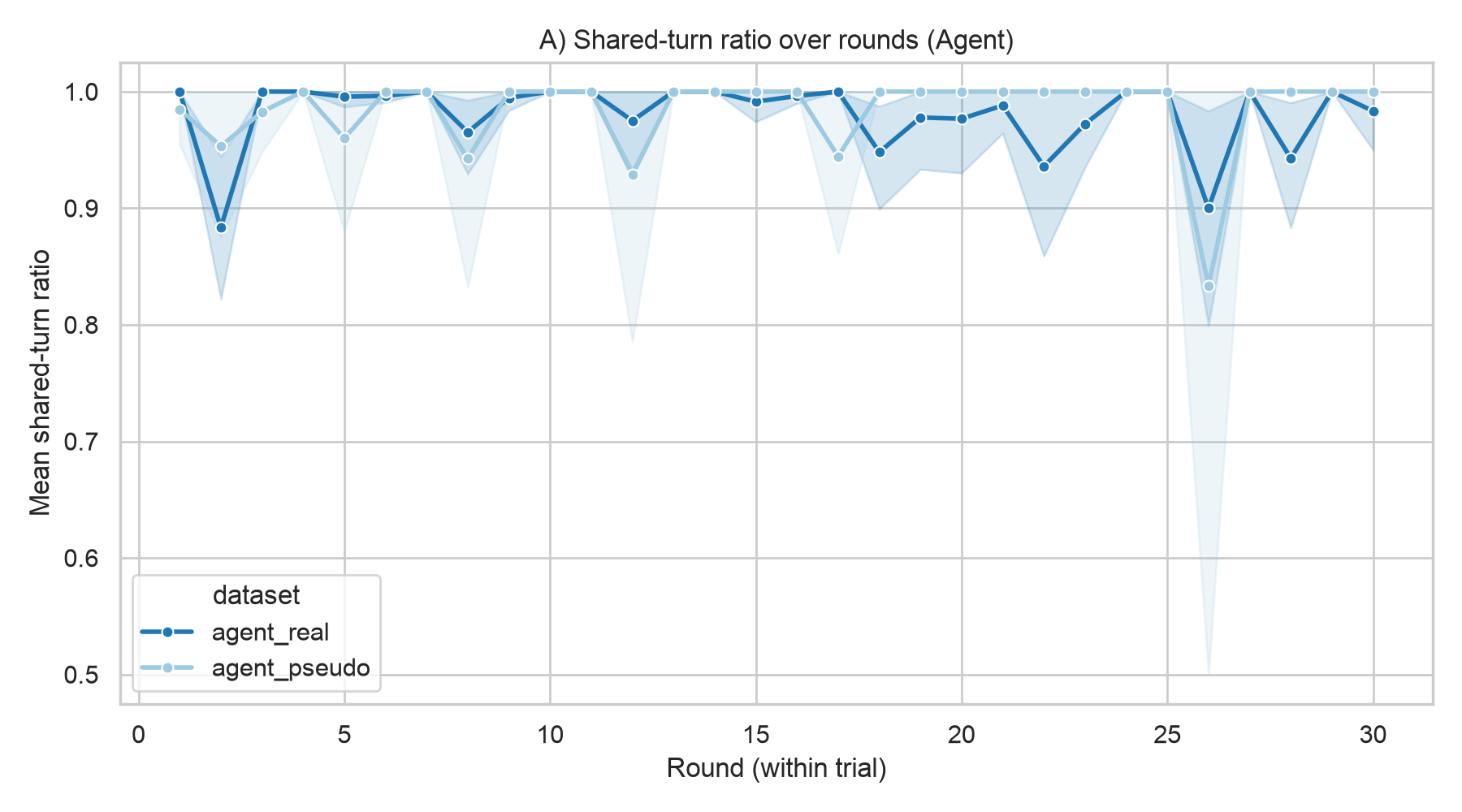}
    \includegraphics[width=\linewidth]{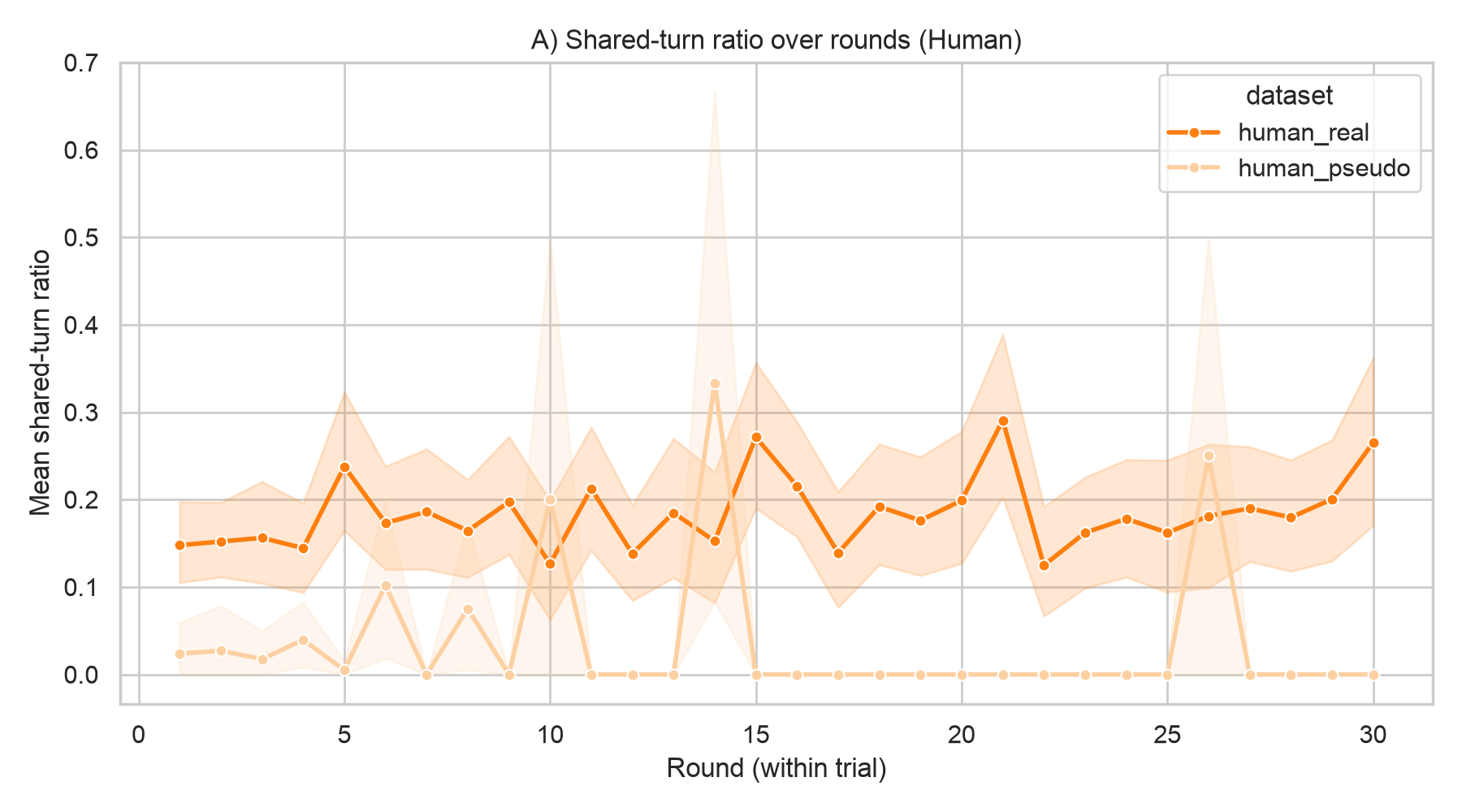}
    \caption{Per-dyad Shared-turn Ratio Shared over Rounds (first 30 rounds shown).}
    \label{fig:esam_shared_turn}
\end{figure}

Figure~\ref{fig:esam_shared_turn} complements the corpus-level turn-ratio analysis in Section~\ref{subsec:micro_alignment} by showing the per-dyad trajectory of the shared-turn ratio over the first 30 rounds, separately for agents (top) and humans (bottom). The agent panel shows that real and pseudo trajectories are tightly overlapping, both hovering at 0.95--1.0 with overlapping confidence bands throughout, with no visible upward trend over rounds. This is consistent with the corpus-level GLMM result
(OR\,$=$\,1.26, 95\% CI $[0.99, 1.61]$): agent overlap is at
ceiling regardless of whether the partner is real or pseudo, and no temporal signature of partner-specific entrainment emerges. The human panel, by contrast, shows a gradual rise in the real trajectory from $\sim$0.10 in the first few rounds toward
$\sim$0.30+ by round~30, while the pseudo trajectory usually remains flat near zero. This temporal divergence
mirrors the corpus-level OR\,$=$\,6.85 effect reported in
Section~\ref{subsec:micro_alignment} and is the per-dyad analogue of the cumulative-entrainment signature predicted by interactive alignment theory: real human dyads build partner-specific reuse incrementally over rounds, whereas pseudo human pairs never do.

\section{Surface-Form Length and Content-word Ratio}

\begin{figure}[tbp]
  \centering
  \includegraphics[width=\linewidth]{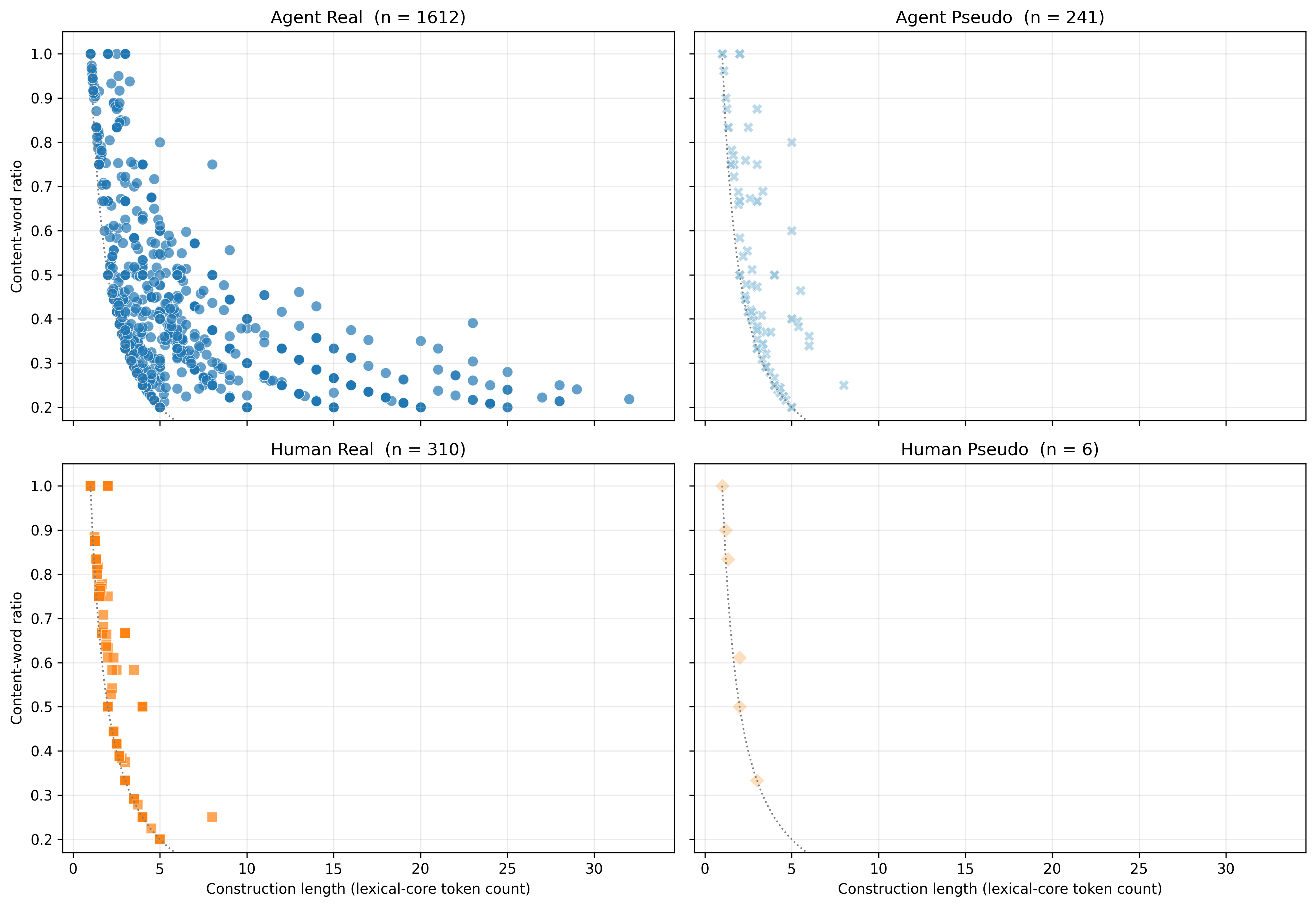}%
  \hfill%
  
  \caption{Lexical core Length vs. Content Ratio.}
  \label{fig:len_con_ratio}
\end{figure}

Figure~\ref{fig:len_con_ratio} plots each shared lexical core by its surface-form token count ($x$-axis) and content-word ratio ($y$-axis). Two distributional differences are visible. First, agent lexical cores span a much wider length range (up to 30+ tokens) than human lexical cores (rarely exceeding 7 tokens), reflecting agent verbosity. Second, human lexical cores cluster tightly in the 1--4 token range at high content ratios, consistent with the compact, noun-phrase-based labels that characterize human entrainment (``the batman,'' ``arms-up guy''). The human distribution peaks where communicative efficiency is highest: short, semantically rich expressions that achieve unambiguous reference with minimal effort.
This distributional contrast maps onto a specific prediction from human reference-game dynamics. \citet{hawkins2020characterizing} show that human convention formation is not only a process of getting shorter but has systematic syntactic structure: closed-class units (determiners, prepositional phrases, relative clauses) drop out in clusters following positive listener feedback, leaving short labels dominated by open-class parts of speech. The tight human cluster in the 1--4 token range at high content-word ratio in Figure~\ref{fig:len_con_ratio} is precisely the open-class residue this process predicts. Agent lexical cores show no such residue: their wide length distribution and variable content ratios indicate that agents regenerate full descriptions rather than incrementally shedding closed-class scaffolding across rounds. The absence of structured syntactic reduction, not only the absence of length reduction, is what distinguishes agent descriptive saturation from human conventionalization.

Agent pseudo-lexical cores overlap substantially with real ones, indicating that lexical core length and composition reflect generation style rather than interactive negotiation. The resulting trade-off is central to the alignment results in Section~\ref{subsec:micro_alignment}: agents are turn-efficient but token-verbose, packing enough discriminative detail into each turn that fewer exchanges suffice, a cost human speakers, who converge on concise partner-specific labels, never pay.

\end{document}